\documentclass[runningheads]{llncs}
\usepackage[T1]{fontenc}
\usepackage{marvosym}
\usepackage{graphicx}
\usepackage{orcidlink}
\usepackage{hyperref}
\hypersetup{hypertex=true,
colorlinks=true,
linkcolor=blue,
anchorcolor=blue,
citecolor=blue}

\usepackage{color}

\usepackage{multirow}
\usepackage{adjustbox}
\usepackage{amsmath}
\usepackage{amssymb}
\usepackage{amsfonts}
\usepackage{float}
\allowdisplaybreaks[4]
\usepackage{booktabs} 
\begin{document}
\title{Rotation-Invariant Completion Network}
%
%

\author{Yu Chen~\orcidlink{0009-0002-2735-504X}\inst{(}\textsuperscript{\Letter}\inst{)}\and
Pengcheng Shi~\orcidlink{0000-0002-9463-6370}
}
\authorrunning{Yu Chen et al.}
%
\institute{Xi’an Jiaotong University, Xi’an, China\\
\email{\{chenyu123,spcbruea\}@stu.xjtu.edu.cn}}
\maketitle              
\begin{abstract}
Real-world point clouds usually suffer from incompleteness and display different poses. While current point cloud completion methods excel in reproducing complete point clouds with consistent poses as seen in the training set, their performance tends to be unsatisfactory when handling point clouds with diverse poses. We propose a network named \textbf{R}otation-\textbf{I}nvariant \textbf{C}ompletion \textbf{Net}work (RICNet), which consists of  two parts: a Dual Pipeline Completion Network (DPCNet) and an enhancing module. Firstly, DPCNet generates a coarse complete point cloud. The feature extraction module of DPCNet can extract consistent features, no matter if the input point cloud has undergone rotation or translation. Subsequently, the enhancing module refines the fine-grained details of the final generated point cloud. RICNet achieves better rotation invariance in feature extraction and incorporates structural relationships in man-made objects. To assess the performance of RICNet and existing methods on point clouds with various poses, we applied random transformations to the point clouds in the MVP dataset and conducted experiments on them. Our experiments demonstrate that RICNet exhibits superior completion performance compared to existing methods.

\keywords{Point cloud completion \and Rotation invariance   \and 3D vision.}
\end{abstract}
\section{Introduction}
Point clouds are widely used for representing the 3D world in computer vision and robotics tasks, but they are often incomplete in real-world scenarios due to limitations of LiDAR scannings, such as object occlusion and range constraints, which result in information loss and constrain point cloud practicality. 

Among the existing point cloud completion methods, \cite{2pcn,3,4ECG,5Cascade} rely on deterministic partial-to-complete mappings to generate complete shapes, lacking the ability to conditionally generate depending on the incomplete point cloud. Additionally, they fail to capture important attributes such as geometric symmetries, regular arrangements, and surface smoothness. To address these issues, VRCNet\cite{7} further incorporates relational structural attributes to enhance the recovery of fine details in point clouds. Although numerous point cloud completion methods proposed in recent years have shown impressive completion performance on various datasets, they only perform well on point clouds with a specific pose in the training dataset, while performance significantly deteriorates on rotated point clouds. However, incomplete point clouds in the real world exhibit various poses, and it is essential for point cloud completion networks to have good generalization performance for point clouds with different poses.

To address these challenges, we propose the \textbf{R}otation-\textbf{I}nvariant \textbf{C}ompletion \textbf{Net}work (RICNet), comprising a Dual Pipeline Completion Network (DPCNet) and an enhancing module. DPCNet generates a coarse complete point cloud, while the enhancing module enhances the detailed features and structural relationships of the coarse point cloud. DPCNet adopts an encoder-decoder architecture. We extract the rotation invariant features by considering the correlation not only between points and their neighbors, but also the internal relationships among the neighboring points. Our feature extraction module ensures consistent predictions during training and testing, regardless of whether the input point cloud has undergone rigid transformations like rotation and translation.  Inspired by Pluralistic image completion\cite{8}, our first module DPCNet uses a dual-path architecture comprising a reconstruction path of the complete point cloud and a completion path of the incomplete point cloud. The reconstruction path adopts a VAE framework, with an encoder that embeds the point cloud into a feature space to obtain an intermediate latent distribution of point cloud features. The Decoder then reconstructs the complete shape from this code. In a similar vein, the completion path attempts to reconstruct a complete shape using the latent distributions and features of the partial input. The encoder and decoder weights are shared between both paths, with the exception of the distribution inference layers. During the training process, we regularize the consistency between the posterior distribution of the encoded incomplete point cloud and the prior distribution of the complete point cloud. Regarding the enhancing module, there exist several plausible options. Inspired by the effectiveness of fine-grained detail and structural relationship enhancement in the coarse point cloud completion of VRCNet\cite{7}, we adopt RENet to infer correlated structures from the incomplete point cloud observations and the generated rough framework after the completion network of the coarse point cloud stage.

We conducted experiments on the MVP dataset. To assess the completion performance and compare it with existing methods on point clouds with different poses, we randomly initialized rigid transformations for each point cloud and trained on them. Then we test the network on both the original and transformed point clouds. Numerous experiments show RICNet's exceptional completion outcomes on both original and transformed point clouds in the MVP dataset.

In conclusion, the following contributions are made by this paper: \textbf{1)} We propose a network architecture for rotation-invariant point cloud completion, consisting of a Dual Pipeline Completion Network (DPCNet) and an enhancing module. \textbf{2)} We have designed a rotation-invariant feature extraction module for learning features of rotated point clouds. This module can be transferred to other point cloud tasks to enhance the network's rotation invariance. \textbf{3)} We compared the completion rates of our approach with other methods on the MVP dataset point clouds after random rotations. The results show that our RICNet outperformed the existing methods.
\section{Related Works}
\subsection{Rotation-Invariant Convolution}
Previous models for point clouds like \cite{9,21} often ignore the rotation invariance. PointNet\cite{1pointnet} only used a T-net to learn the rotational features of point clouds, but its performance suffered greatly from the simple object rotations if there is no data augmentation. The disorderliness of the point cloud makes it difficult to capture the rotational features. For instance, if some rotation is applied to the point cloud in PointNet++\cite{9}, the segmentation performance becomes particularly poor. \cite{11} proposed a simpler approach using hand-crafted features based on Euclidean distance and angle. However, the resulting local features may not be sufficiently expressive, leading to reduced accuracy. \cite{12} proposes a global context-aware convolution approach that utilizes anchor points and local reference frames (LRF). \cite{13,14} further utilized LRF to learn rotation-invariant local descriptors to improve performance. \cite{15} proposed an effective framework based on several features. However, it will be unstable in the presence of noise or outliers, limiting overall performance. To tackle these difficulties, RIConv++\cite{16} proposed a straightforward and powerful convolutional operator. This operator is specifically designed to capture robust rotation-invariant features from local regions. It enhances feature distinctiveness by considering the correlation not only between the points and their neighboring points, but also the internal relationships among the neighboring points.
To address these challenges, RIConv++\cite{16} proposed a simple and effective convolutional operator, which is designed to extract robust rotation-invariant features from local regions and improve feature discrimination by considering the relationships between interest points and their neighbors, as well as the internal relationships among neighbors.
\subsection{Point Cloud Completion}
PCN \cite{2pcn} employs folding operations for upsampling to generate coarse completion by leveraging global features learned from incomplete point clouds. The decoder proposed by TopNet \cite{6TopNet} can effectively predicts complete shapes. \cite{4ECG,5Cascade,11} utilize local features to enhance the quality of their completion results. The aforementioned approaches focus on the generation of overall shape frameworks while neglecting fine-grained details. Moreover, they largely learn deterministic mappings from partial to complete, ignoring structural relationships in man-made objects. Slice Sequential Network\cite{20} utilizes a slice-based approach to process incomplete point clouds. By slicing the input point cloud, it can effectively capture the inter-slice information and generates the missing parts with high fidelity. VRCNet\cite{7} proposed a variational framework consisting of PMNet and RENet. PMNet leverages the advantages of explicit reconstruction and generative modeling to generate coarse and complete point clouds. RENet further extracts relational point features through point self-attention and point selection modules to refine local shape details. The recently proposed methods have achieved impressive completion results on various point cloud datasets. However, they lack rotation invariance and generalization ability for handling rotated point clouds, leading to unsatisfactory results for point clouds with diverse poses. The neglect of rotation invariance in current point cloud completion networks restricts their practical applicability in real-world scenarios.
\section{Our Method}
\begin{figure}[t]
\centering
\includegraphics[width=\textwidth]{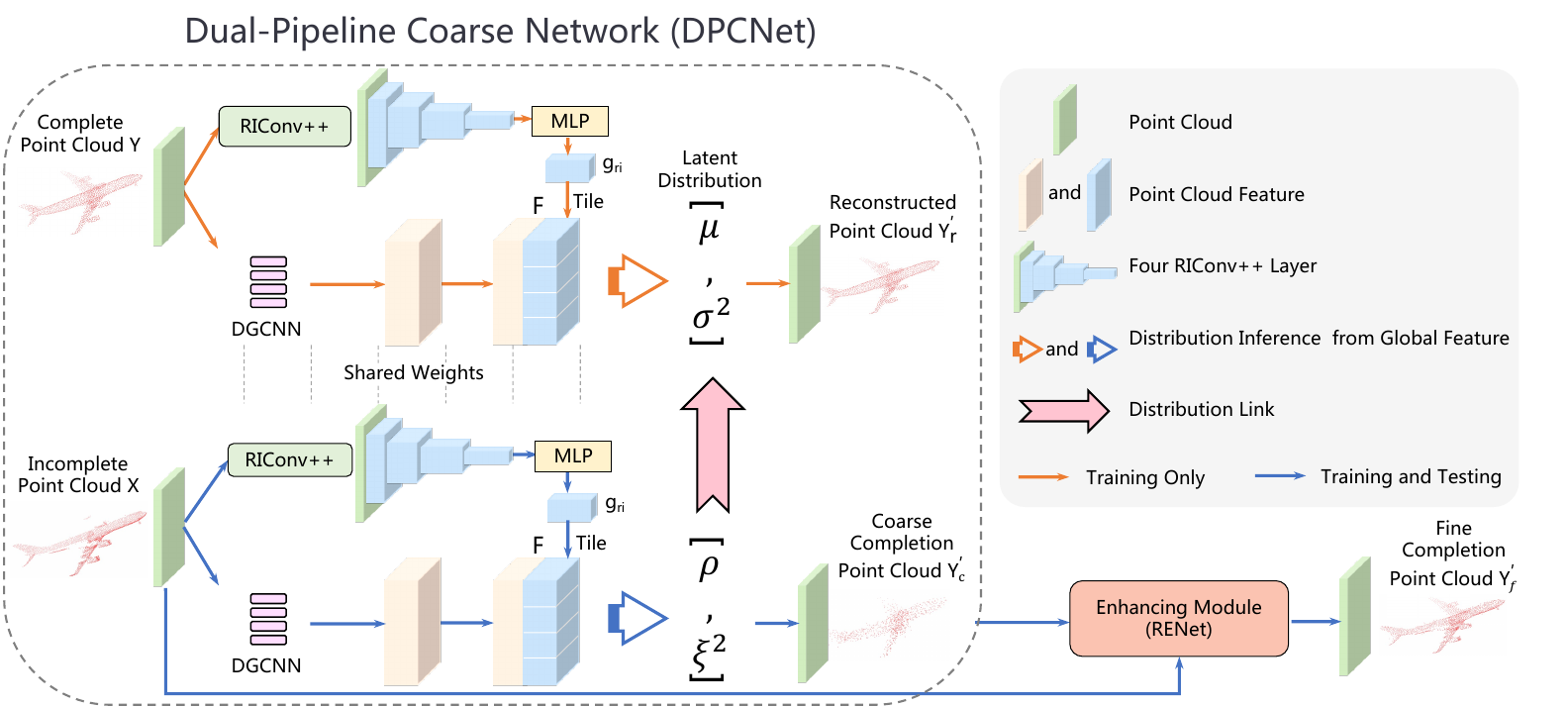}
\caption{\textbf{The architecture of RICNet. }DPCNet generates a coarse complete point cloud  \({Y_{c}}'\), while the enhancing module refines the fine-grained details of the final output  to produce the final completion \({Y_{f}}'\). DPCNet employs an encoder-decoder architecture, with the two parallel paths sharing weights. The distribution link ensures consistency between the posterior distribution of the encoded partial point cloud and the prior distribution of the complete point cloud.} 
\label{fig1}
\end{figure}

To address the issue of unsatisfactory results in point clouds with varied poses, we propose a network architecture for rotation-invariant point cloud completion, as shown in Fig. \ref{fig1}. We define \(X\) as a partial observation, while \(Y\) represents the ground truth. The objective of our network is to generate a predicted complete point cloud \({Y_{f}}'\) based on the provided input \(X\).
\subsection{Rotation-Invariant Embedding Module\label{Section 3.1}}
We employ RIConv++ to explicitly extract global features, while simultaneously integrating DGCNN\cite{17} to enhance the extraction of local features. 
The RIConv++ network, comprising four convolutional layers, processes the input point cloud to obtain a global feature \(g_{ri}\). The definition of the convolutional layer can be found in \nameref{Section 3.1.2}. Besides, DGCNN converts each point into a point feature vector \(f_{i}\). We connect \(g_{ri}\) to each \(f_{i}\) to create a point feature matrix \(F\) where each row represents a concatenated feature vector \([g_{ri}, f_{i}]\). Extensive experiments demonstrate that our feature extraction module inherits the rotation invariance and global feature extraction capabilities from RIConv++, while also effectively extracting local features with the assistance of DGCNN.

\subsubsection{Rotation-Invariant Convolution\label{Section 3.1.2}}
\begin{figure}[t]
\centering
\includegraphics[width=\textwidth]{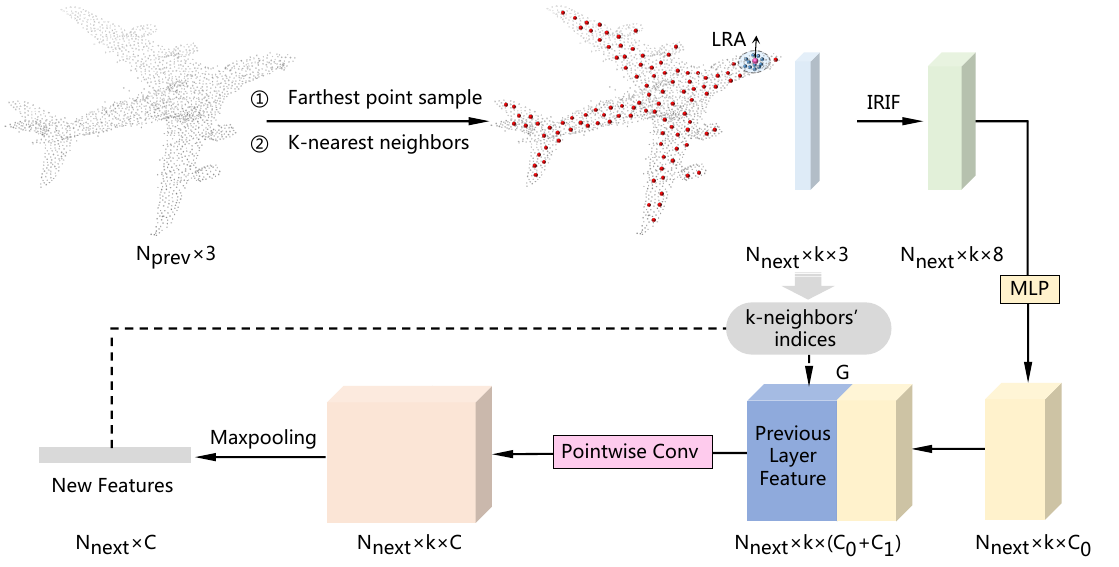}
\caption{\textbf{A Rotation-Invariant Convolutional Layer.} We perform farthest point sampling to select a set of reference points \( \left \{  r_{j} \right \} \) (red dots) from the point cloud. K-NN is utilized to obtain local point sets \( X_{j}=\left \{x_{n}\right \} \) (blue dots) for each reference point \(r_j\). IRIFs are then calculated and transformed into a high-dimensional space using an MLP. These features are combined with previous layer features (if any), followed by pointwise convolution and Maxpooling to generate the New Features.} \label{fig2}
\end{figure}

RIConv++ employs a sequence of four convolutional layers to explicitly extract rotation-invariant global features of point clouds, denoted as \(g_{ri}\). Fig. \ref{fig2} illustrates the structure of a rotation-invariant convolutional layer. The reference point set obtained in a convolutional layer will serve as the input point cloud for the subsequent layer, and the New Features obtained in the current layer will be combined with the point cloud features of the next layer, generating the feature matrix \(G\) for the subsequent layer, as indicated by the dashed line in Fig. \ref{fig2}.

\begin{figure}[H]
\centering
    \includegraphics[width=0.4\textwidth]{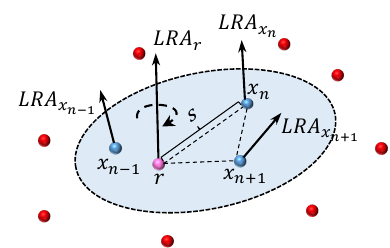}
\caption{\textbf{IRIF Construction.} Given a reference point r, the K-NN algorithm retrieves k nearest neighbors  \( X_{j}=\left \{x_{n}\right \} \) and orders the points clockwise. The IRIF of a neighbor point \(x_{n}\) is represented by (\ref{eq2}). }\label{fig3}
\end{figure}

IRIF serves as a powerful feature representation. Fig. \ref{fig3} can facilitate our comprehension of IRIF, which is a transformation that maps each neighboring point \(x_n\) to a tuple, comprising seven attributes:
\begin{equation}\label{eq2}
t_{n}= [s,\delta ,a_1,a_2,a_3,b_1,b_2,b_3],
\end{equation}
where \(t_n\)  represents the IRIF feature tuples of \(x_n\). \(s\), \(a_1\), \(a_2\), and \(a_3\)  quantify the correlation between neighbor point \(x_{n} \) and the reference point \(r\) (radial direction). \( \delta, b_1, b_2\), and \(b_3\) encode the correlation between \( x_n \) and its adjacent neighbor \(x_{n+1}\) :

\begin{equation}
    \begin{aligned}
        &s =\left \|x_n-r  \right \|,\\ 
        &a_1 =arccos(LRA_{r},\overrightarrow{x_{n}r}),\\
        &a_2 =arccos(LRA_{x_n},\overrightarrow{x_{n}r}),\\
        &a_3 =k_{a}\cdot  arccos(LRA_{x_{n}}, LRA_{r}).\\
        &\delta =arccos(\overrightarrow{x_{n+1}r},\overrightarrow{x_{n}r}),\\
        &b_1 =arccos(LRA_{x_{n}}, \overrightarrow{x_{n}x_{n+1}}),\\
        &b_2 =arccos(LRA_{x_{n+1}}, \overrightarrow{x_{n}x_{n+1}}),\\
        &b_3 =k_{b}\cdot  arccos(LRA_{x_{n}}, LRA_{x_{n+1}}).
    \end{aligned}
\end{equation}

Here, \(LRA\) is a reliable and stable shape descriptor reference vector that is invariant to rotation, as introduced in \cite{10}.  \(LRA_{p_{i}} \) can be represented by normal vectors at point \(p_{i}\). The \(arccos\) values are only defined in the interval \(\left [ 0,\pi  \right ]\), introducing a signed ambiguity. To capture angle and directional information between two vectors, we employ the utilization of signed angles as indicated in Equation (\ref{eq4}).
\begin{equation}
k_{a} =
\left\{
             \begin{array}{lr}
            +1, if\;a_{1}\le a_{2} &  \\
            -1, otherwise &  
             \end{array}
\right.
,
k_{b} =
\left\{
             \begin{array}{lr}
            +1, if\;b_{1}\le b_{2} &  \\
            -1, otherwise &  
             \end{array}
\right.
.
\label{eq4}
\end{equation}
\vspace{-0.75cm}
\vspace{-0.5cm}
After MLP transformation, we obtain a  \(N_{next}\times k\times C_0 \) feature matrix. The previous convolutional layer's features represented as a \(N_{prev}\times C_1 \) matrix will be transmitted to this layer. For each reference point \(r_{j}\),  \(C_1\)-dimensional features are assigned to its k neighboring points based on the k-neighbors' indices. Therefore, we obtain a \(N_{next}\times k\times C_1 \) feature matrix, which will be combined with the features obtained in the current layer to form the feature matrix \(G\). We order the points clockwise by projecting each point \(x_{n}\) onto the local tangent disk. The ordering is established by selecting a starting point (e.g. \(x_{0}\)) and setting its projection \(\overrightarrow{x_{0}p} \) as the reference point on the disk. Then, we compute the angles between the projected points \(\overrightarrow{x_{n}p} \) and \(\overrightarrow{x_{0}p} \) and sort them in ascending order from 0 to 360 degrees. In our implementation, \(x_{0}\) is chosen as the farthest point from the reference point \(p\) in the local neighborhood. Then we apply a 1D convolution to the feature tensor, followed by Maxpooling along the k-nearest neighbors dimension, to obtain the New Feature for this layer. 

In more formulaic terms, we define a point set \(\mho = \left \{ p_i \right \}\) where \(p_i\) denotes the 3D coordinates of the point \(i\), and \(t_i\) represents the IRIF feature of \(p_i\). The convolution operation \(h\) for learning the features of \(\mho\) is expressed as follows:
\begin{equation}
    h(\mho)=\sigma (\kappa ({MLP(t_i): \forall i})),
\end{equation}
Here, \(\sigma\) is an activation function, and \(\kappa  \) is a Maxpooling function. It consists of a 1D convolutional kernel and an ordering function, ensuring rotational invariance is preserved. We denote the IRIF features after passing through the MLP as \(m_i\).
\begin{equation}
    \kappa ({m_i})=Maxpool(1Dconv\left ( order\left ( m_i \right ) \right )),
\end{equation}
Here, the function \(order\) arranges the points in clockwise order by projecting \(p_i\) onto the local tangent disk, and \(1Dconv\) refers to a one-dimensional convolution.
\subsection{DPCNet: Dual-Pipeline Completion Network}

To construct the overall framework of the complete point cloud, we introduce a dual-path design. Fig. \ref{fig1} shows that the architecture consists of a reconstruction path dedicated to the ground truth \(Y\) and a completion path for the incomplete point cloud \(X\). The reconstruction path utilizes a VAE framework, with an encoder (\nameref{Section 3.1}) that embeds the point cloud into a feature space to obtain an intermediate latent distribution of point cloud features. The Decoder then reconstructs the complete shape from this code.

Specifically, the reconstruction path first extracts point cloud features to obtain a feature matrix \(F\). MLP and Maxooling are then applied to derive the global feature \(v_{r}\), from which we extract the  latent distributions \(\lambda (v_{r}|Y) \) for the complete shape \(Y\), and then use a decoding  distribution to recover a complete shape \({Y_{r}}'\). The reconstruction path is only used for training. The loss function of the reconstruction path is formulated as:
\begin{equation}\label{eq6}
    L_{rec}=-KL[\lambda(v_{r}|Y)\left |  \right |  p(v_{r})]+L_{CD}\left ( {Y_{r} }' ,Y \right )
\end{equation}
Here,  \(KL \) represents the Kullback-Leibler divergence, \(p(v_r)=N(0,I)\) is a pre-defined Gaussian conditional prior, and CD denotes the symmetric chamfer distance (CD) loss formulated as:
\begin{equation}\label{eq9}
    L_{CD}(P,Q)=\frac{1}{\left | P \right | } \sum_{x\in P}\min_{y\in Q}\left \| x-y \right \|^{2}+\frac{1}{\left | Q \right | } \sum _{y\in Q}\min_{x\in P}\left \| x-y \right \|   ^{2}. 
\end{equation}
Here, \(x\) and \(y\) stand for points that belong to point clouds \(P\) and \(Q\).

The structure of the completion path is similar to that of the reconstruction path and shares its encoder and decoder weights, except for the distribution inference layer. Its main objective is to reconstruct the complete shape \({Y_{c}}'\) based on the global feature \(v_c\) and the latent distribution \( \varphi (v_c| X) \) obtained from the input \(X\). To effectively use the most significant features from the incomplete point cloud, a learned conditional distribution \( \lambda (v_r| Y) \), which encodes its corresponding complete 3D shape \(Y\), is employed to adjust the latent distribution \(\varphi  \) during training. Thus, \( \lambda (v_r| Y) \) forms the prior distribution, and \( \varphi (v_c| X) \) serves as the posterior importance sampling function. The completion path is characterized by the following loss function formulation:
\begin{equation}\label{eq7}
    L_{com}= -KL[\lambda (v_r|Y)|| \varphi (v_c| X)] +L_{CD}\left ( {Y_{c}}',Y  \right )
\end{equation}
\subsection{Enhancing Module}
Inspired by the effectiveness of fine-grained detail and structural relationship enhancement in the coarse point cloud completion of VRCNet\cite{7}, we utilize the Relation Enhancement Network (RENet) as our enhancement module to improve structural relationships and recover the fine completion \({Y_{f}}'\)  based on the initial completion  \({Y_{c}}'\). The R-PSK module, comprising an MLP and PSK module, plays a crucial role in RENet. The PSK module enables the adaptive fusion of structure relations learned at various scales, allowing neurons to dynamically adjust their receptive field sizes to match the distinct scales of PSA's relation structures. PSA effectively aggregates point features in local neighborhoods by leveraging learned relations. The loss function for the enhancing module is formulated as:
\begin{equation}\label{eq8}
    L_{fine}=L_{CD}\left ( {Y_{f}}',Y  \right )
\end{equation}
\subsection{Loss Function}
The joint training loss \(L\) of our RICNet is formulated as follows:
\begin{equation}
L=\omega _{rec}L_{rec}+\omega _{com}L_{com}+\omega _{fine}L_{fine}
\end{equation}
Here, \(L_{rec} \), \(L_{com}\) and \( L_{fine}\) are the losses defined in (\ref{eq6}), (\ref{eq7}), and (\ref{eq8}). \(\omega _{rec}\),  \(\omega_{com}\), and \(\omega_{fine}\) are weighted parameters. 
\section{Experiments}
We utilized PyTorch to implement our network and our model was trained on an NVIDIA TITAN Xp GPU. We employ the Adam optimizer \cite{18}, with an initial learning rate of \(1e^{-4}\), decaying by 0.7 every 40 epochs. The Chamfer distance (\ref{eq9}) is used to calculate the distance between the final predicted complete point cloud (Y') generated by the network  and the ground truth point cloud (Y), while the F-score \cite{19} is utilized to measure the distance between object surfaces.

To evaluate the proposed network's completion performance under different point cloud poses, we randomly apply a rigid transformation to each point cloud and train the network on these transformed point clouds, testing it on both the original and transformed point clouds. 

\vspace{-0.5cm}
\begin{table}[H]
	\begin{minipage}[t]{0.48\textwidth}
		\centering
  	\caption{Completion results of the original cloud without transformation. Our RICNet outperforms all  existing methods.}	\label{tab1}
		\begin{tabular}{c|cc}
    \hline
    Method         & $CD\downarrow$    &$F1\uparrow$     \\ \hline \hline
    TopNet\cite{6TopNet}     & 29.46 & 0.167 \\
    PCN\cite{2pcn}        & 26.13 & 0.173 \\
    Cascade\cite{5Cascade}    & 21.48 & 0.309 \\
    MSN\cite{3}        & 22.79 & 0.344 \\
    ECG\cite{4ECG}        & 13.48 & 0.424 \\
    VRCNet\cite{7}     & 10.58 & 0.461 \\ \hline
   \textbf{ RICNet(Ours)} &\textbf{ 9.14 } & \textbf{0.470 }\\ \hline 
    \end{tabular}
	\end{minipage}
 	\hspace{.15in}
	\begin{minipage}[t]{0.48\textwidth}
		\centering
		\caption{Completion results of the transformed cloud. Our RICNet also outperforms all  existing methods.}	\label{tab2}
		\begin{tabular}{c|cc}
    \hline
    Method           & $CD\downarrow$      & $F1\uparrow$     \\ \hline \hline
    TopNet\cite{6TopNet}     & 20.561 & 0.193 \\
    PCN\cite{2pcn}         & 15.68  & 0.236 \\
    Cascade\cite{5Cascade}    & 16.46  & 0.318 \\
    MSN\cite{3}        & 15.81  & 0.362 \\
    ECG\cite{4ECG}        & 9.44   & 0.440 \\
    VRCNet \cite{7}    & 7.75   & 0.469 \\ \hline
    \textbf{RICNet(Ours) }& \textbf{7.57   }&\textbf{ 0.477} \\ \hline
    \end{tabular}
	\end{minipage}
\end{table}
\vspace{-0.5cm}
\vspace{-0.5cm}
\begin{table}[H]
\renewcommand\arraystretch{1.5}
\caption{Completion results on the transformed cloud of MVP. Here we adopt the CD loss (multiplied by \(10^4\))  as the evaluation metric. Lower values indicate better performance. RICNet stands out as the top-performing model overall. }\label{tab3}
\resizebox{1.0\linewidth}{!}{
\begin{tabular}{c|cccccccccccccccc|c}
\hline
\textbf{Method}     & Airplane & Cabinet & Car    & Chair  & Lamp   & Sofa   & Table  & Watercraft & Bed    & Bench  & Bookshelf & Bus    & Guitar & Motorbike & Pistol & Skateboard & \textbf{Avg. }  \\
\hline \hline
PCN\cite{2pcn}        & 6.556    & 22.618  & 11.042 & 19. 017 & 22.916 & 17.612 & 18.55  & 13.197     & 27.329 & 13.012 & 21.363    & 15.028 & 4.482  & 9.956     & 8.831  & 7.157      & 14.916 \\
TopNet\cite{6TopNet}      & 8.986    & 31.152  & 14.174 & 25.553 & 29.211 & 24.133 & 23.326 & 14.861     & 34.320  & 16.196 & 28.045    & 20.389 & 6.506  & 11.706    & 10.966 & 9.350       & 19.304 \\
MSN\cite{3}         & 5.678    & 25.246  & 13.853 & 18.138 & 21.559 & 19.218 & 16.064 & 13.624     & 25.646 & 10.483 & 20.125    & 17.017 & 4.546  & 10.122    & 9.210   & 8.391      & 14.932 \\
Cascade\cite{5Cascade}     & 6.435    & 29.065  & 11.847 & 19.728 & 20.418 & 19.819 & 20.836 & 11.205     & 26.301 & 12.92  & 24.034    & 17.094 & 4.453  & 8.942     & 8.364  & 6.659      & 15.507 \\
ECG\cite{4ECG}         & 3.522    & 15.009  & 7.631  & 11.191 & 12.196 & 11.754 & 10.326 & 8.463      & 18.117 & 6.888  & 13.469    & 9.327  & 2.409  & 5.858     & 5.515  & 4.504      & 8.510   \\
VRCNet\cite{7}      & 2.679    & 13.904  & \textbf{7.108}  &\textbf{ 8.932}  & 8.031  & 9.644  & \textbf{8.629}  & 6.572      & 14.939 & 5.287  & \textbf{10.386 }   & 7.451  & 1.816  &\textbf{ 4.713}     & 4.004  & \textbf{2.959}      & 7.315  \\ \hline
\textbf{RICNet(Ours)} & \textbf{2.644}    & \textbf{12.950}   & \underline{7.607}  &  \underline{9.096}  & \textbf{7.341}  & \textbf{9.277}  & \underline{8.774}  &\textbf{ 6.115}      & \textbf{13.771} & \textbf{4.950}   & \underline{11.582}    & \textbf{7.134}  & \textbf{1.669}  & \underline{4.859}     & \textbf{3.929 } & \underline{2.975}      & \textbf{7.167} \\ \hline
\end{tabular}
}
\end{table}
\vspace{-0.5cm}
\vspace{-0.5cm}
\begin{table}[H]
\renewcommand\arraystretch{1.5}
\caption{Completion results on the transformed cloud of MVP. Here we adopt the F-score \cite{20}  as the evaluation metric. Higher values indicate better performance. RICNet also stands out as the top-performing model overall. } \label{tab4}
\resizebox{1.0\linewidth}{!}{
    \begin{tabular}{c|cccccccccccccccc|c}
    \hline
    \textbf{Method}     & Airplane & Cabinet & Car   & Chair & Lamp  & Sofa  & Table & Watercraft & Bed   & Bench & Bookshelf & Bus   & Guitar  & Motorbike & Pistol & Skateboard & \textbf{Avg}.  \\ \hline \hline
    PCN\cite{2pcn}         & 0.509    & 0.102   & 0.153 & 0.143 & 0.211 & 0.116 & 0.192 & 0.267      & 0.108 & 0.278 & 0.120      & 0.197 & 0.519   & 0.243     & 0.346  & 0.470       & 0.248 \\
    TopNet\cite{6TopNet}      & 0.409    & 0.078   & 0.131 & 0.120  & 0.170  & 0.098 & 0.170  & 0.243      & 0.092 & 0.245 & 0.097     & 0.150  & 0.464   & 0.226     & 0.309  & 0.400        & 0.212 \\
    MSN\cite{3}         & 0.650     & 0.193   & 0.239 & 0.302 & 0.440  & 0.23  & 0.337 & 0.408      & 0.249 & 0.446 & 0.238     & 0.289 & 0.640    & 0.375     & 0.439  & 0.494      & 0.373 \\
    Cascade\cite{5Cascade}     & 0.615    & 0.145   & 0.209 & 0.230  & 0.392 & 0.179 & 0.252 & 0.394      & 0.197 & 0.362 & 0.171     & 0.268 & 0 .639 & 0.326     & 0.436  & 0.538      & 0.294 \\
    ECG\cite{4ECG}         & 0.737    & 0.242   & 0.300   & 0.375 & 0.547 & 0.283 & 0.403 & 0.495      & 0.308 & 0.556 & 0.289     & 0.391 & 0.748   & 0.442     & 0.541  & 0.656      & 0.382 \\
    VRCNet\cite{7}      & \textbf{0.780}     & \textbf{0.250}    & 0.308 & 0.403 & 0.619 & 0.302 & 0.430  & 0.530       & 0.333 & 0.606 & 0.315     & 0.405 & \textbf{0.824}   & 0.474     & 0.579  & 0.700        & 0.491 \\ \hline
    \textbf{RICNet(Ours)} & \underline{0.772}    &\underline{0.246}   & \textbf{0.310 } & \textbf{0.414} & \textbf{0.636} & \textbf{0.308} & \textbf{0.439} & \textbf{0.541}      & \textbf{0.346} & \textbf{0.617} & \textbf{0.318}     & \textbf{0.406} & \underline{0.814}   & \textbf{0.486}     & \textbf{0.579}  & \textbf{0.706}      & \textbf{0.496} \\ \hline
    \end{tabular}
}
\end{table}
\vspace{-0.5cm}
\vspace{-0.5cm}
\subsubsection{Quantitative Evaluation\label{Section 4.1}}
The completion results on the original point cloud are shown in Table \ref{tab1}, while the results on randomly rotated point clouds are presented        in Table \ref{tab2}. Previous methods only consider point clouds with fixed poses, and notable performance degradation can be observed when applied to rotated point clouds. RICNet performs better than existing methods on both non-rotated and randomly rotated  point clouds.

MVP dataset comprises 16 categories of point clouds. Table \ref{tab3}, \ref{tab4} quantitatively the completion results on incomplete point clouds with different poses among all categories. Although RICNet may have slightly inferior performance in certain specific categories compared to VRCNet, it still stands out as the top-performing model overall. 

\begin{table}[H]
\caption{Qualitative completion results on the transformed cloud using different methods. The point clouds in the dataset consist of 2,048 points. RICNet surpasses VRCNet in capturing the overall structure in completion and exhibits significantly superior completion capability compared to other existing methods.}\label{tab5}
\resizebox{1.0\linewidth}{!}{
\begin{tabular}{@{}ccccccccc@{}}
Input & PCN\cite{2pcn}  & TopNet\cite{6TopNet} & MSN\cite{3} & Cascade\cite{5Cascade} & ECG\cite{4ECG} & VRCNet\cite{7} & \begin{tabular}[c]{@{}l@{}}RICNet\\ (Ours)\end{tabular} & \begin{tabular}[c]{@{}l@{}}Ground\\ Truth\end{tabular} \\ 
\specialrule{0em}{5pt}{5pt}
        \includegraphics[width=0.1111\textwidth]{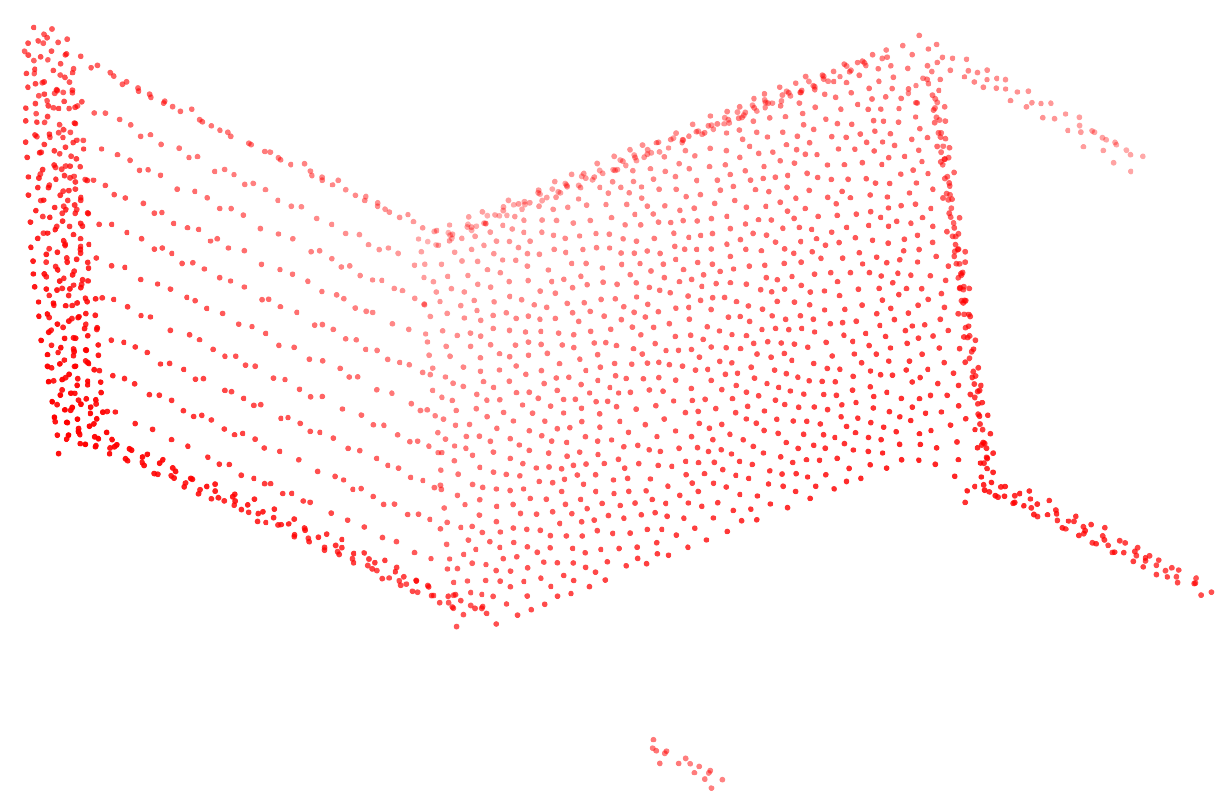}
      & \includegraphics[width=0.1111\textwidth]{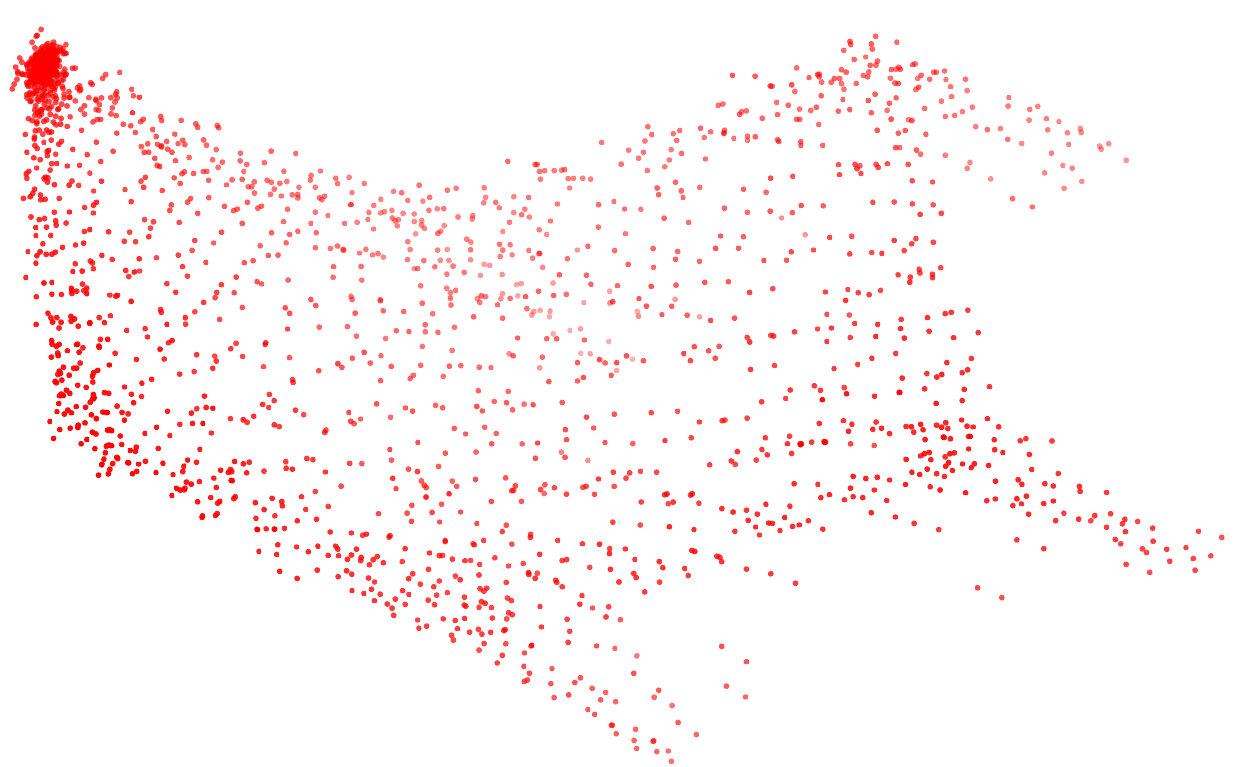}
      & \includegraphics[width=0.1111\textwidth]{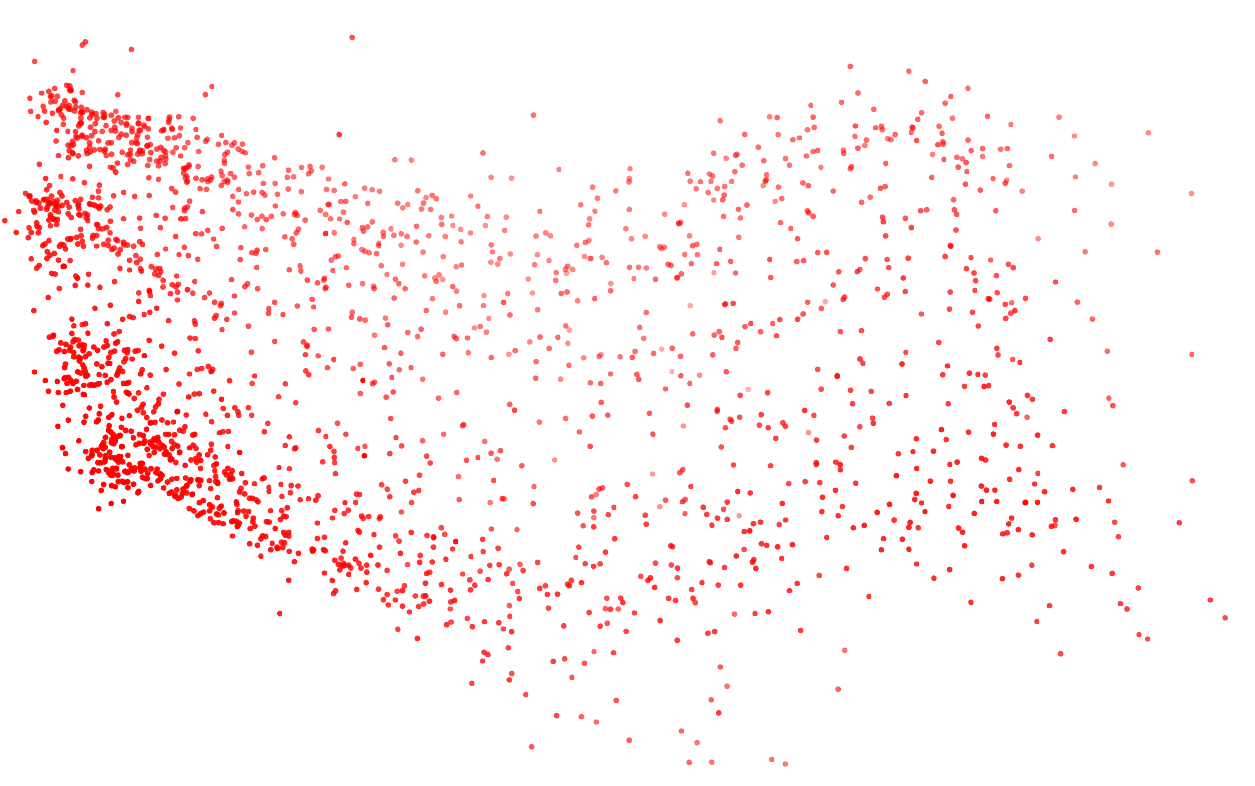}
      & \includegraphics[width=0.1111\textwidth]{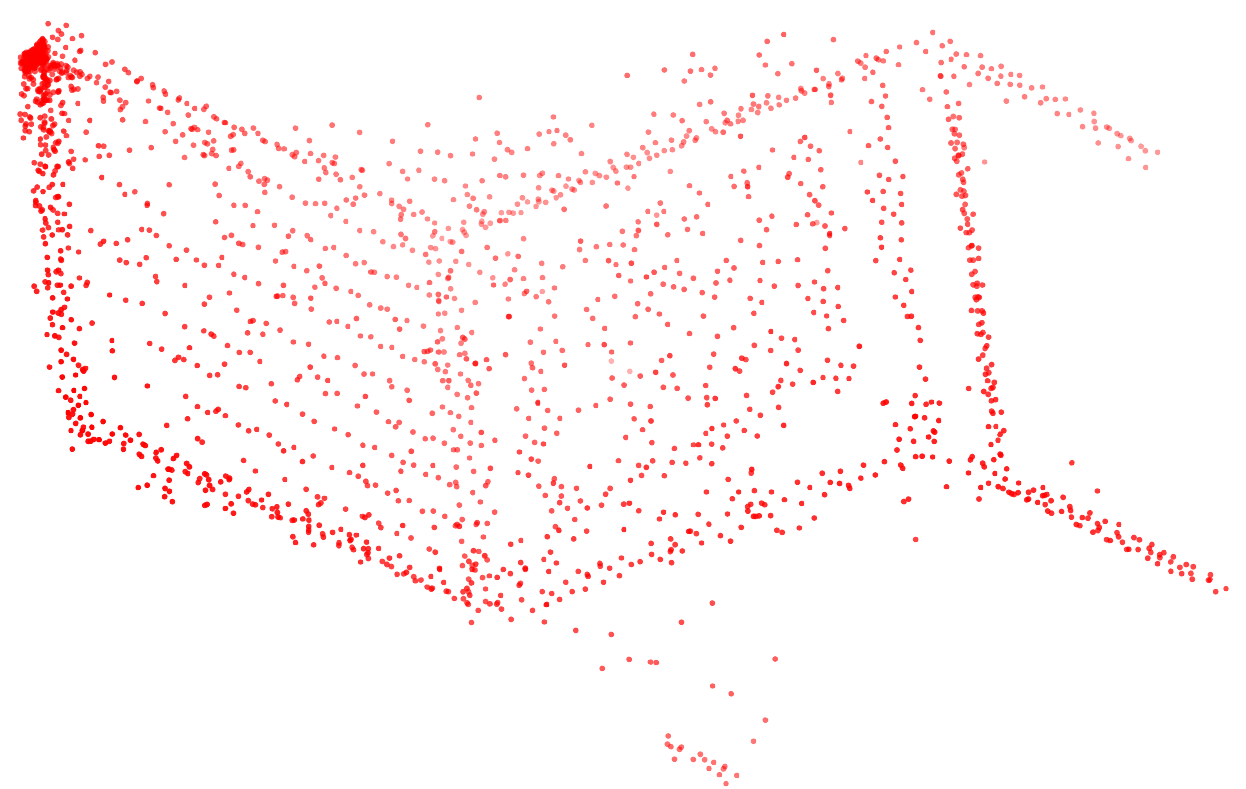}        
      & \includegraphics[width=0.1111\textwidth]{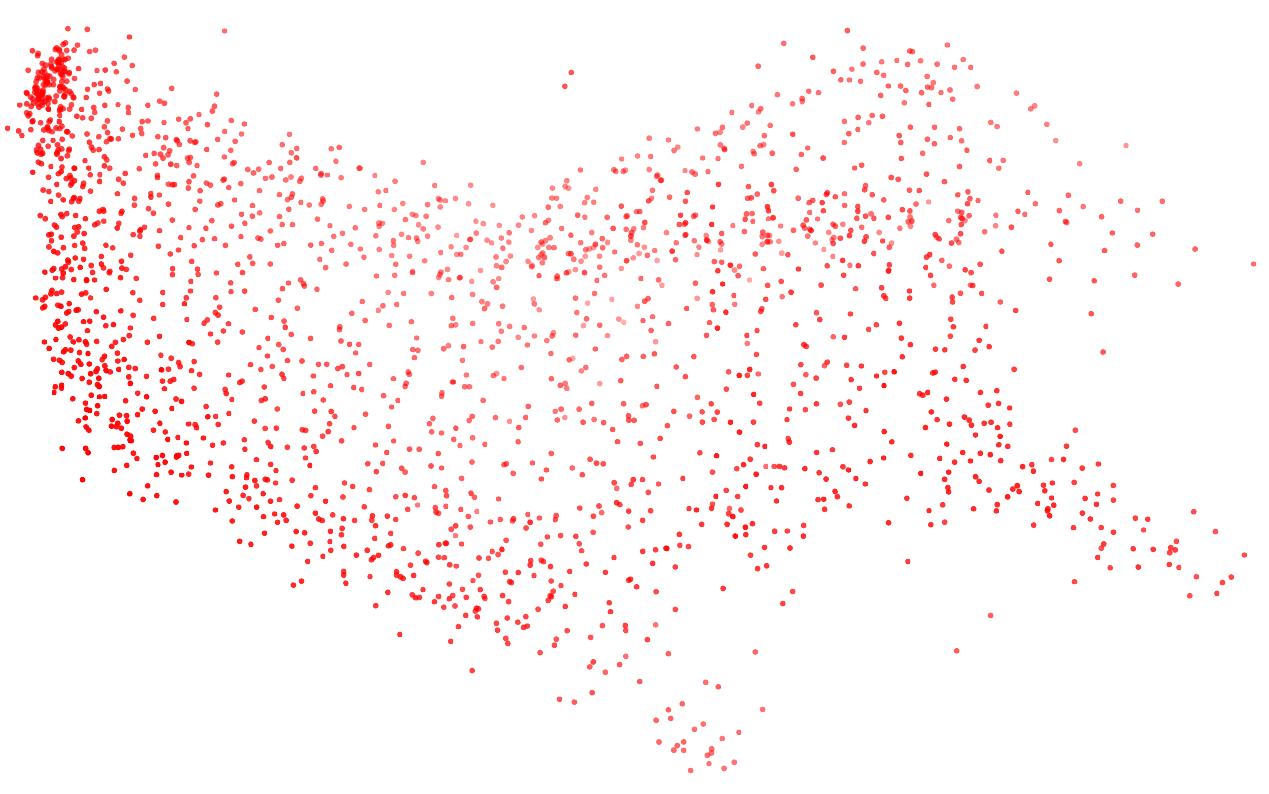}    
      & \includegraphics[width=0.1111\textwidth]{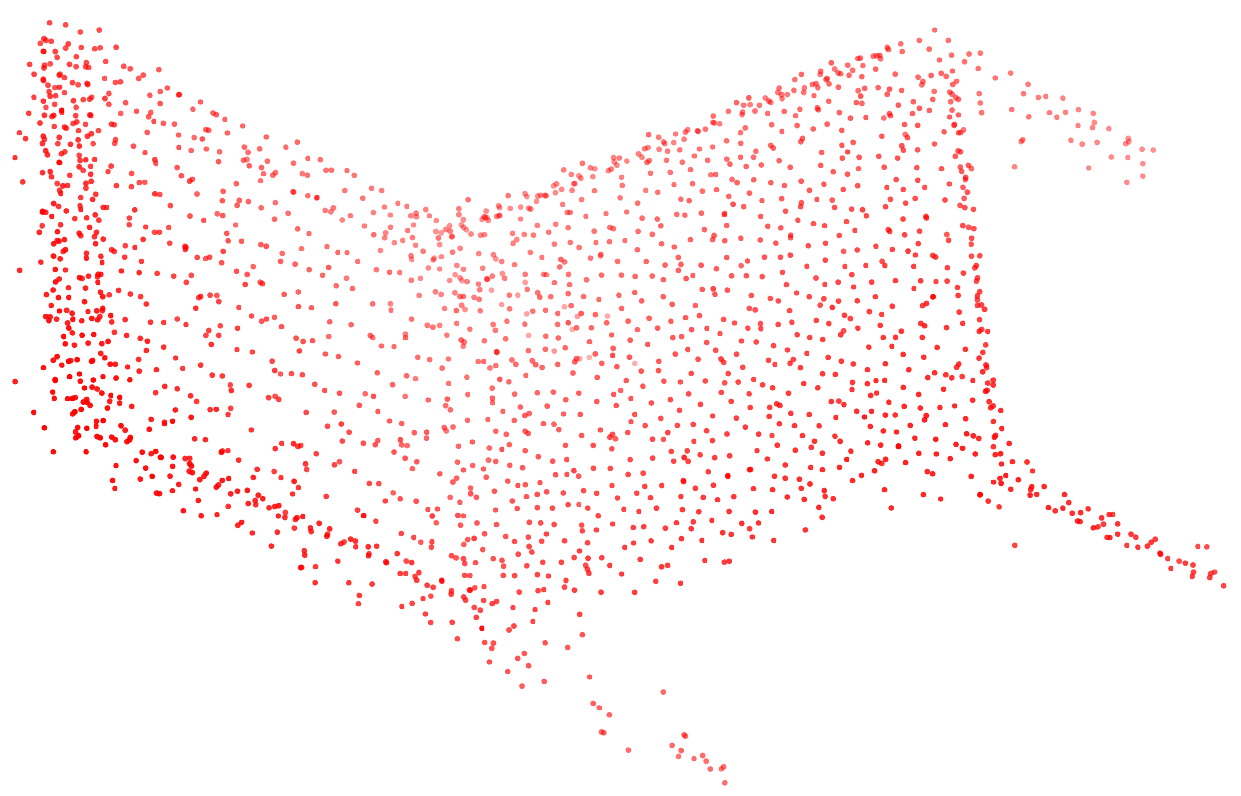}      
      & \includegraphics[width=0.1111\textwidth]{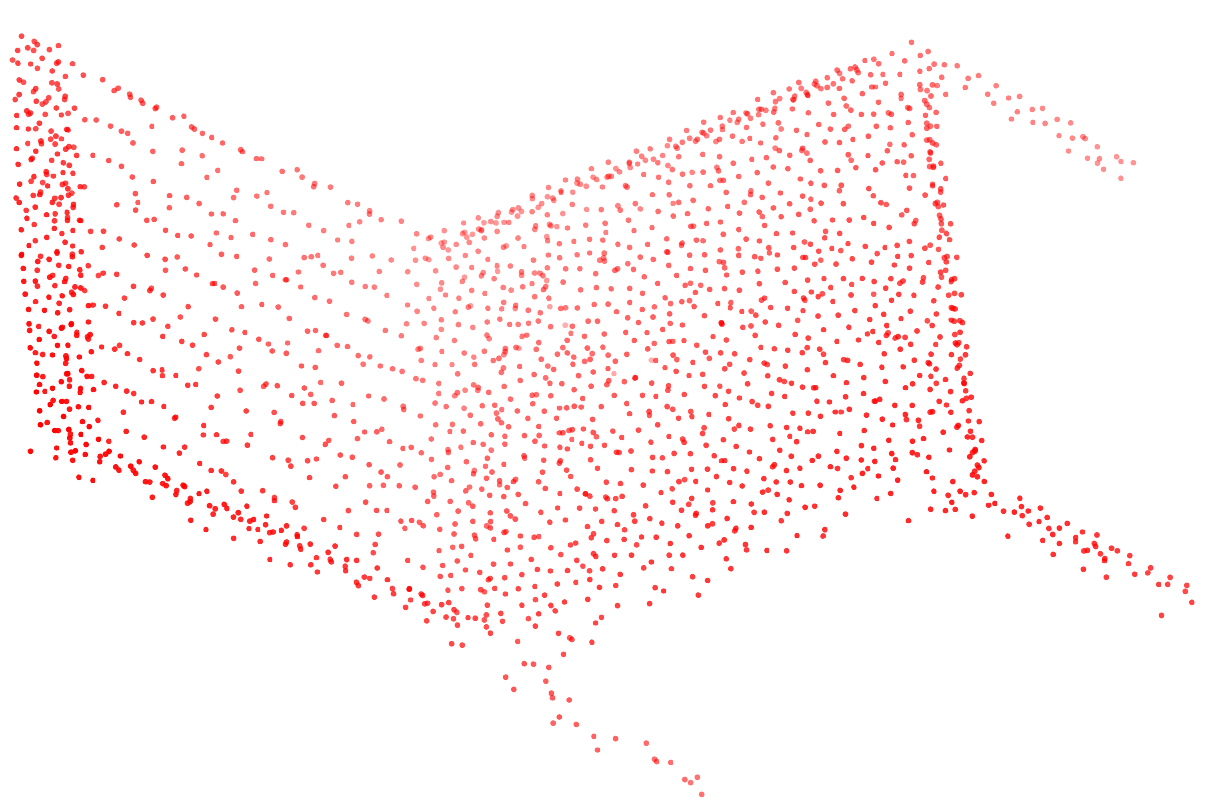}              
      & \includegraphics[width=0.1111\textwidth]{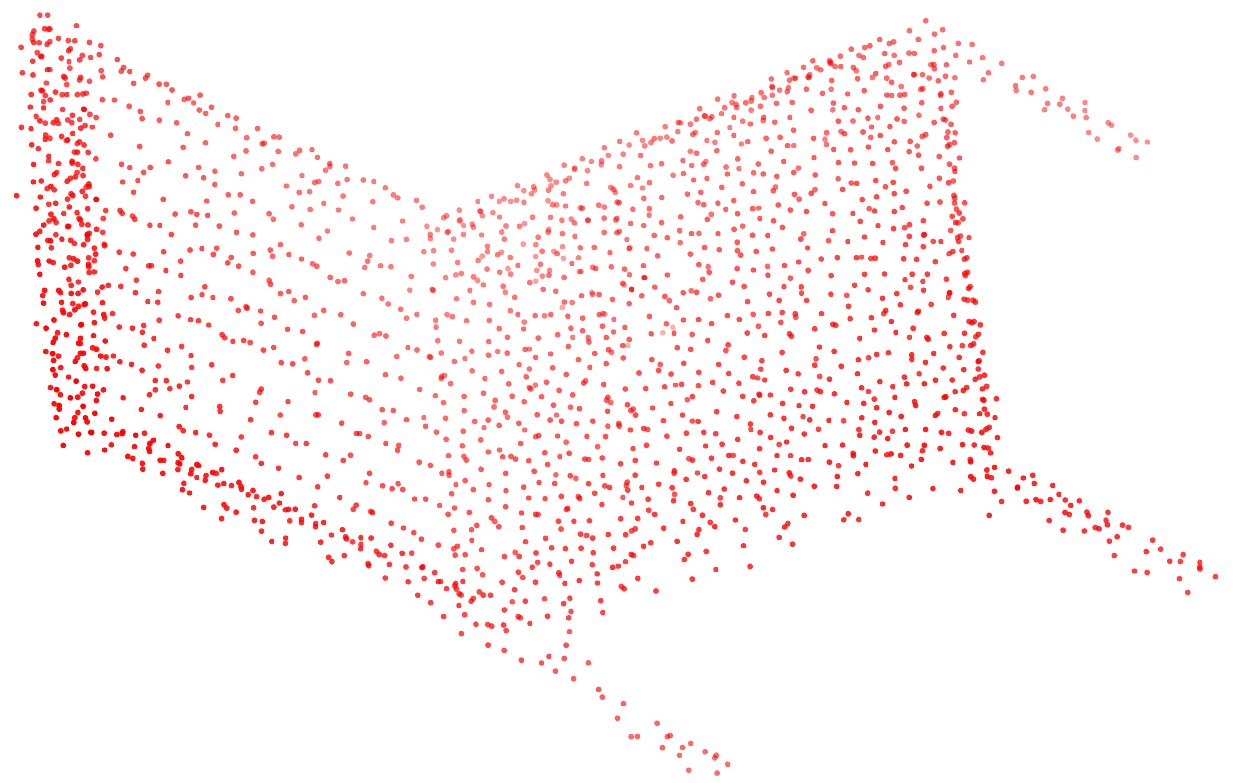}
      & \includegraphics[width=0.1111\textwidth]{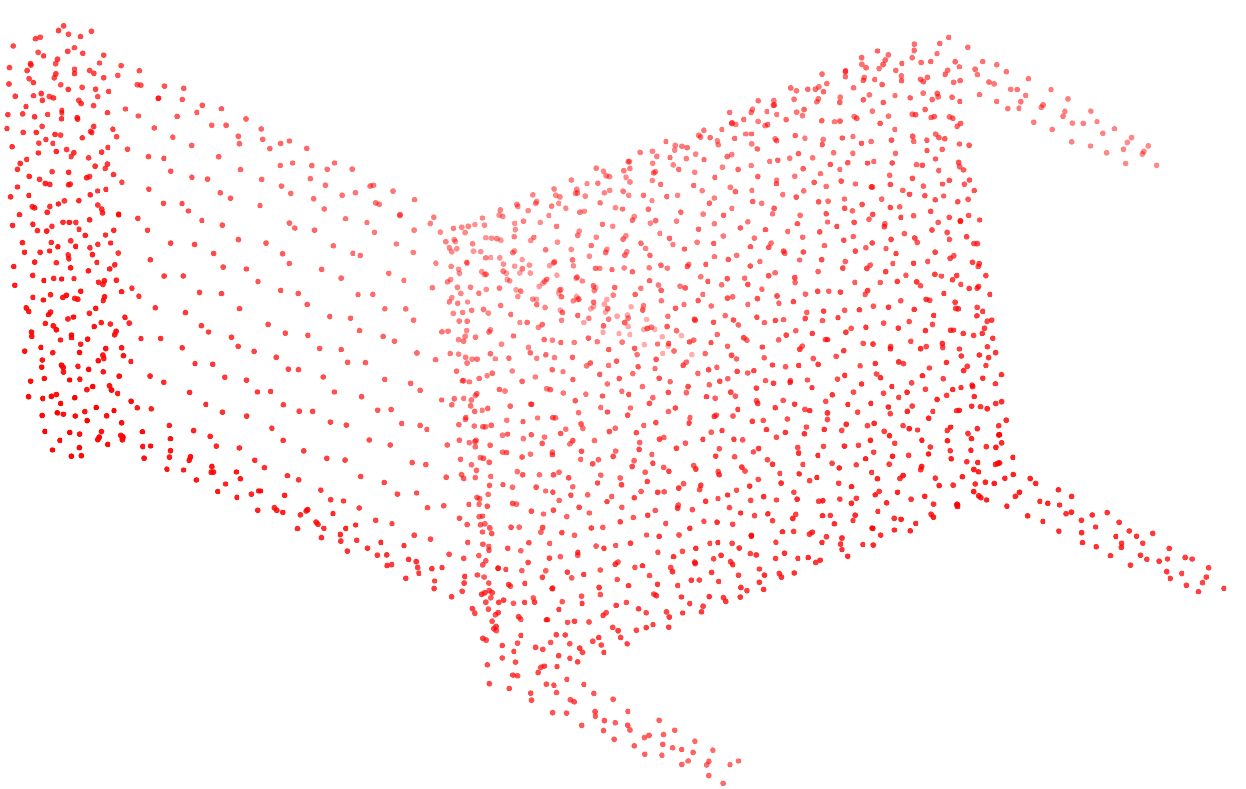} \\
      \specialrule{0em}{0pt}{5pt}
         \includegraphics[width=0.1111\textwidth]{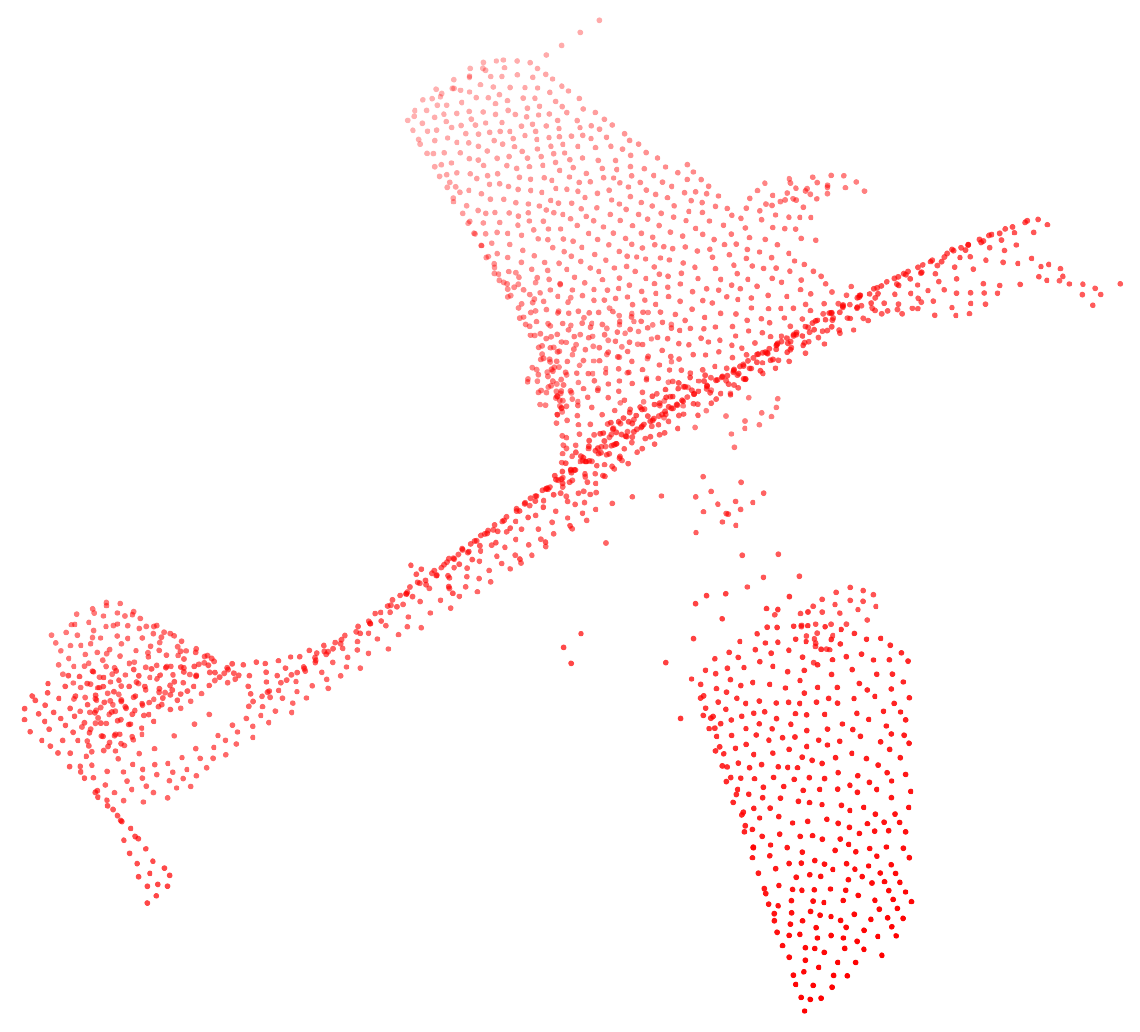}
      & \includegraphics[width=0.1111\textwidth]{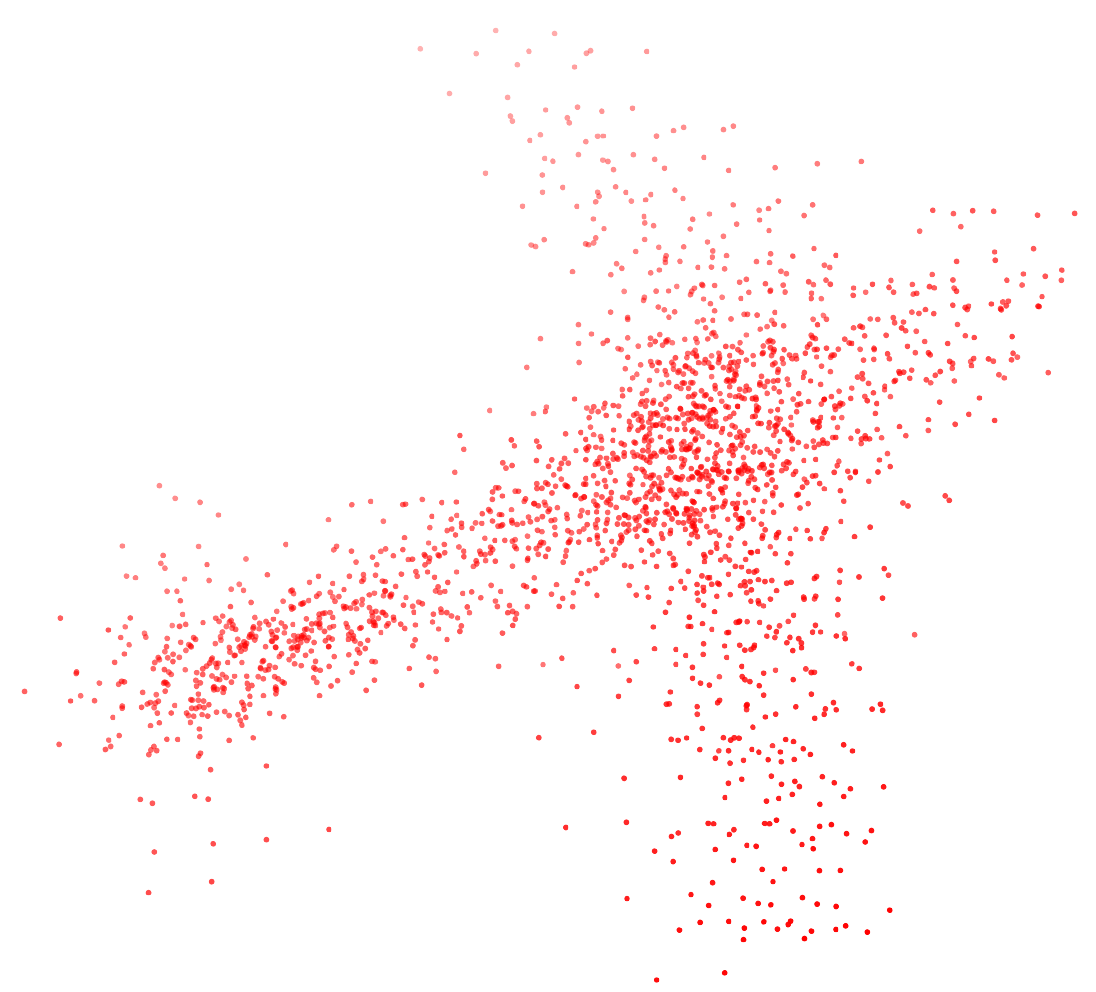}
      & \includegraphics[width=0.1111\textwidth]{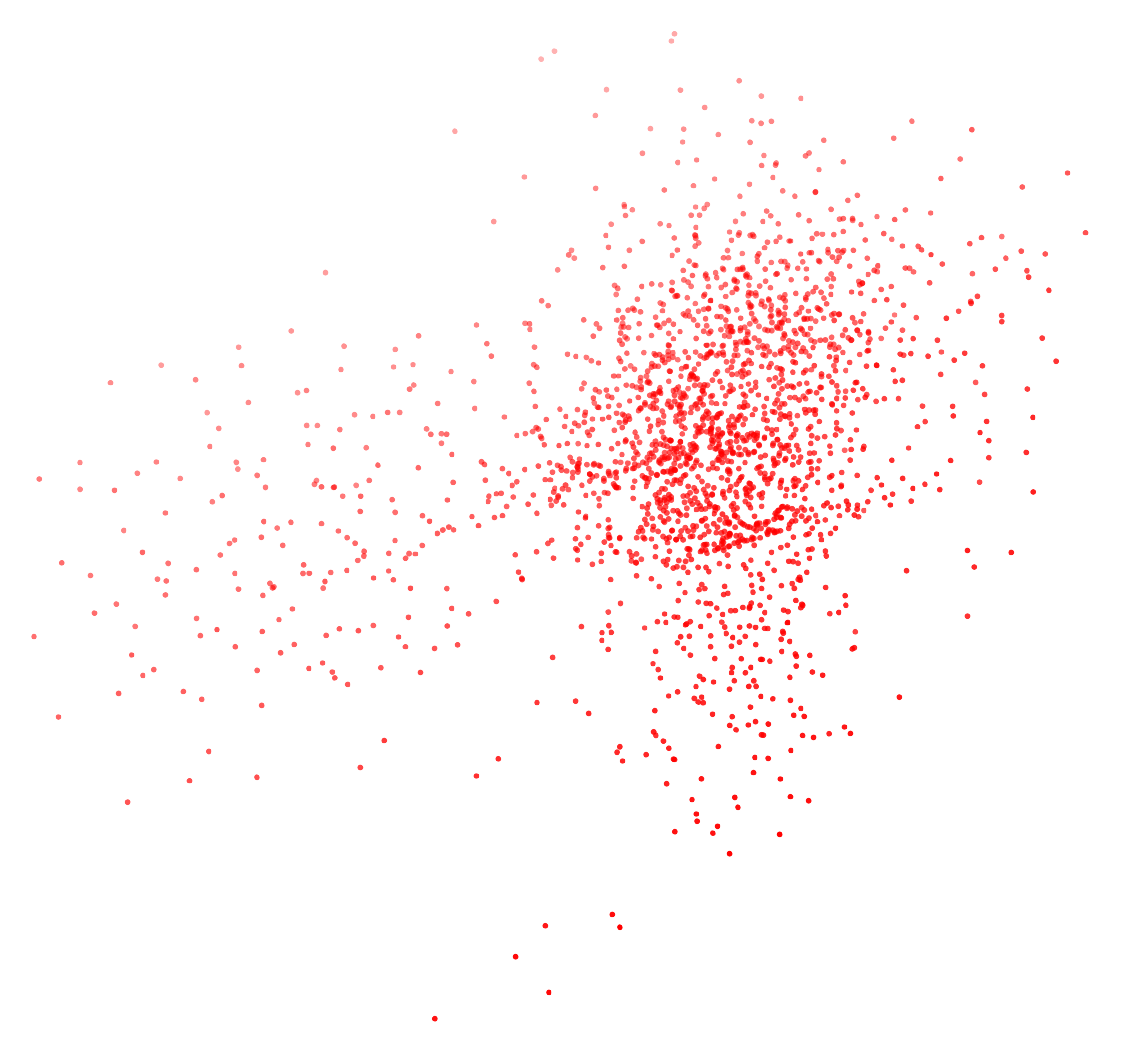}
      & \includegraphics[width=0.1111\textwidth]{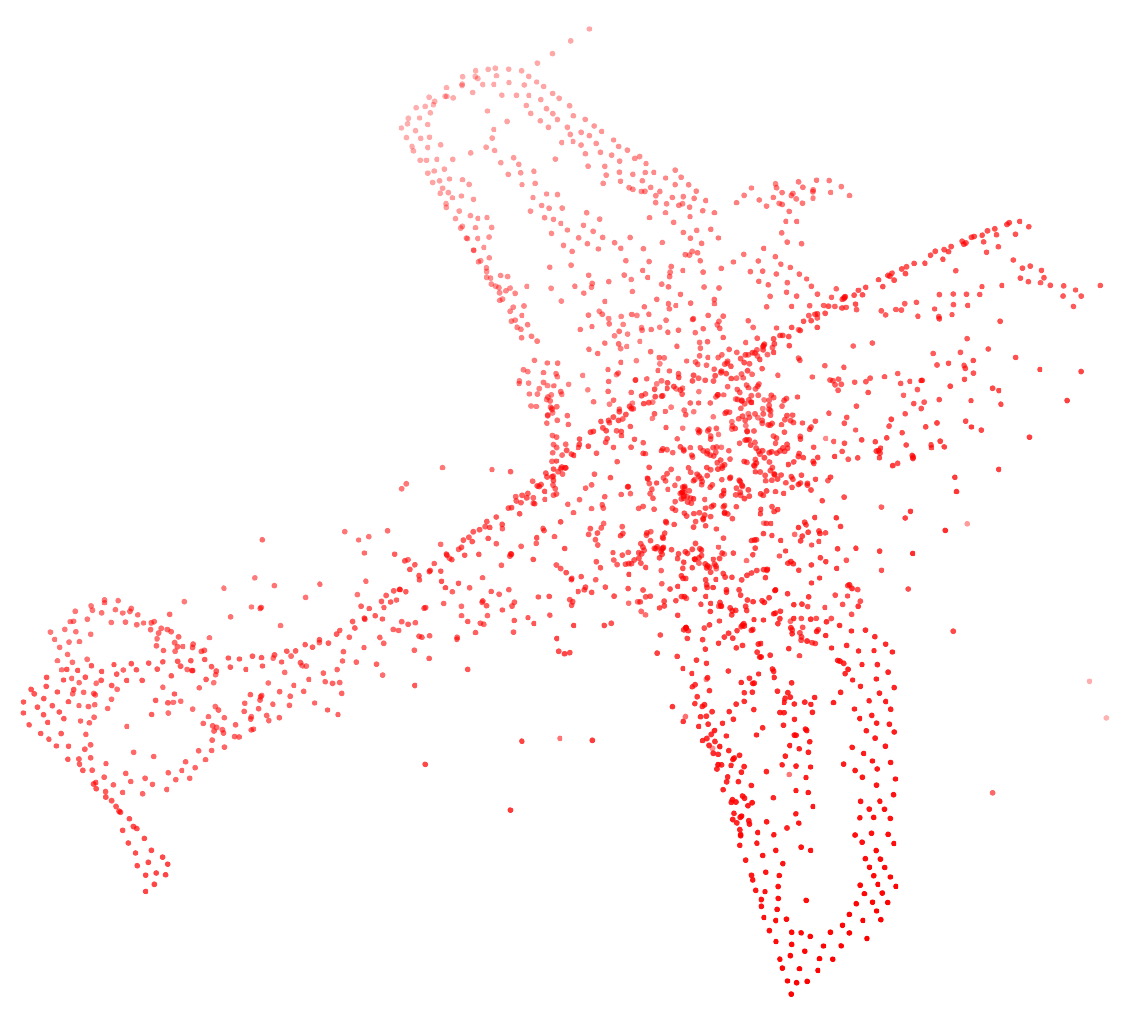}        
      & \includegraphics[width=0.1111\textwidth]{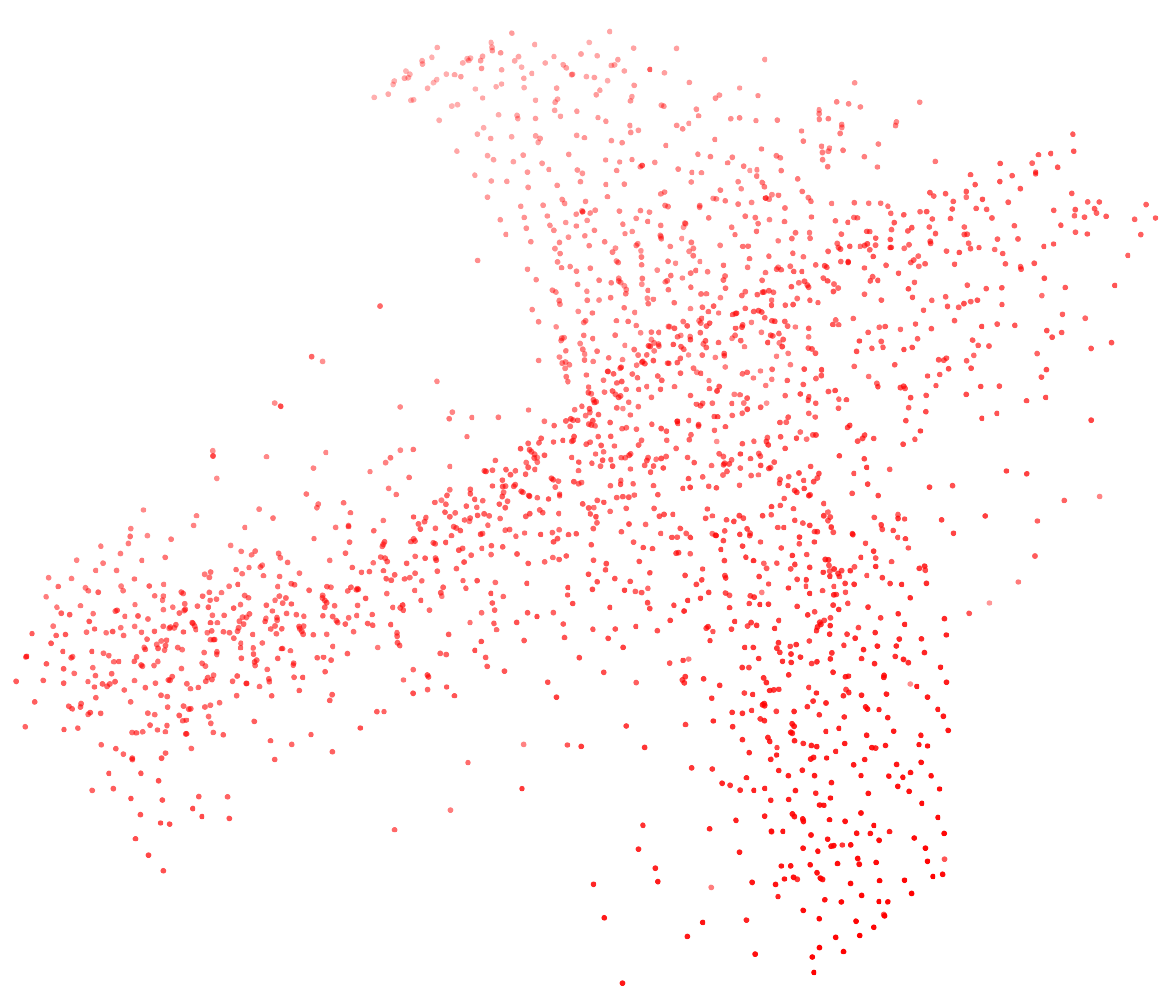}    
      & \includegraphics[width=0.1111\textwidth]{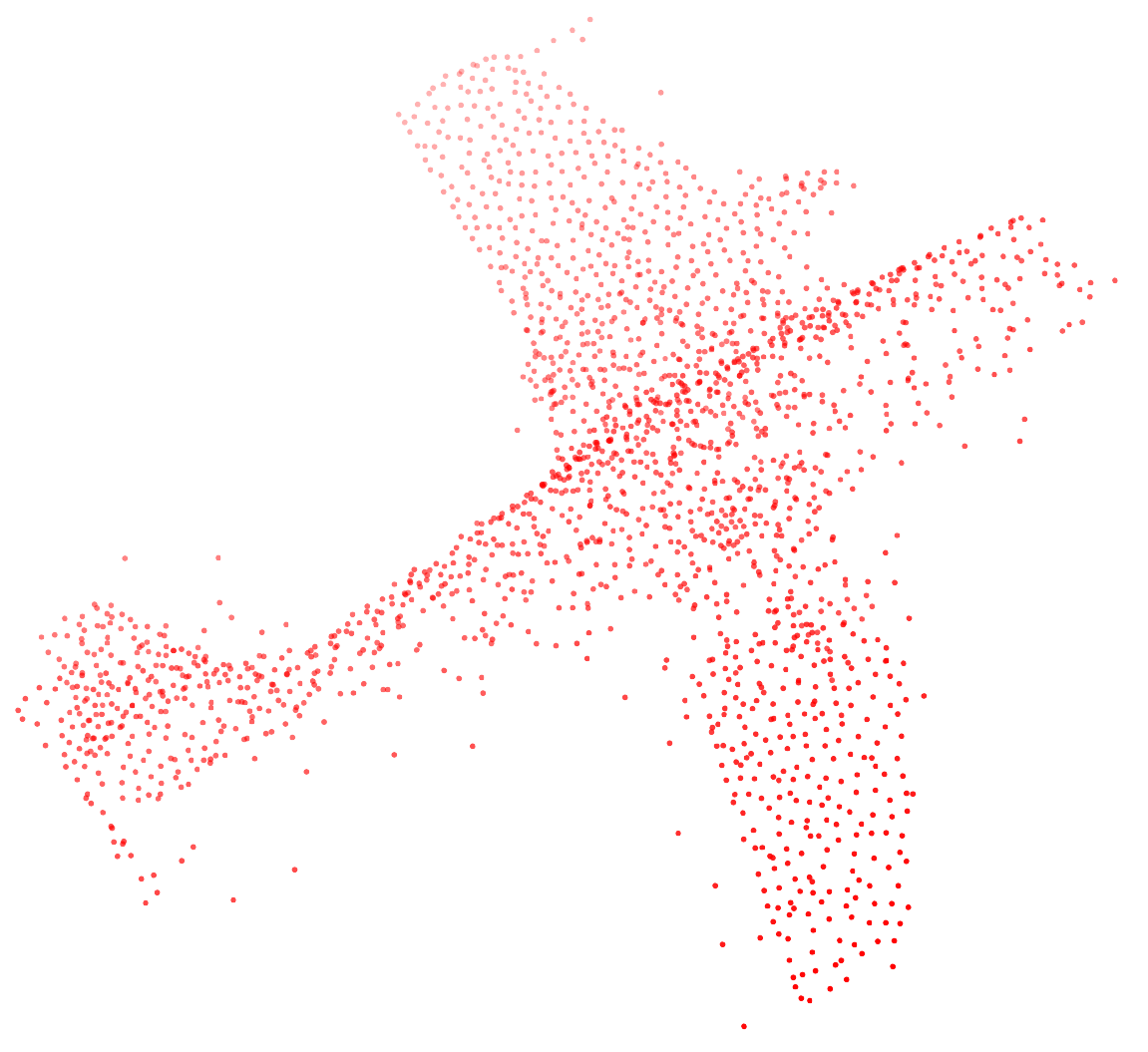}      
      & \includegraphics[width=0.1111\textwidth]{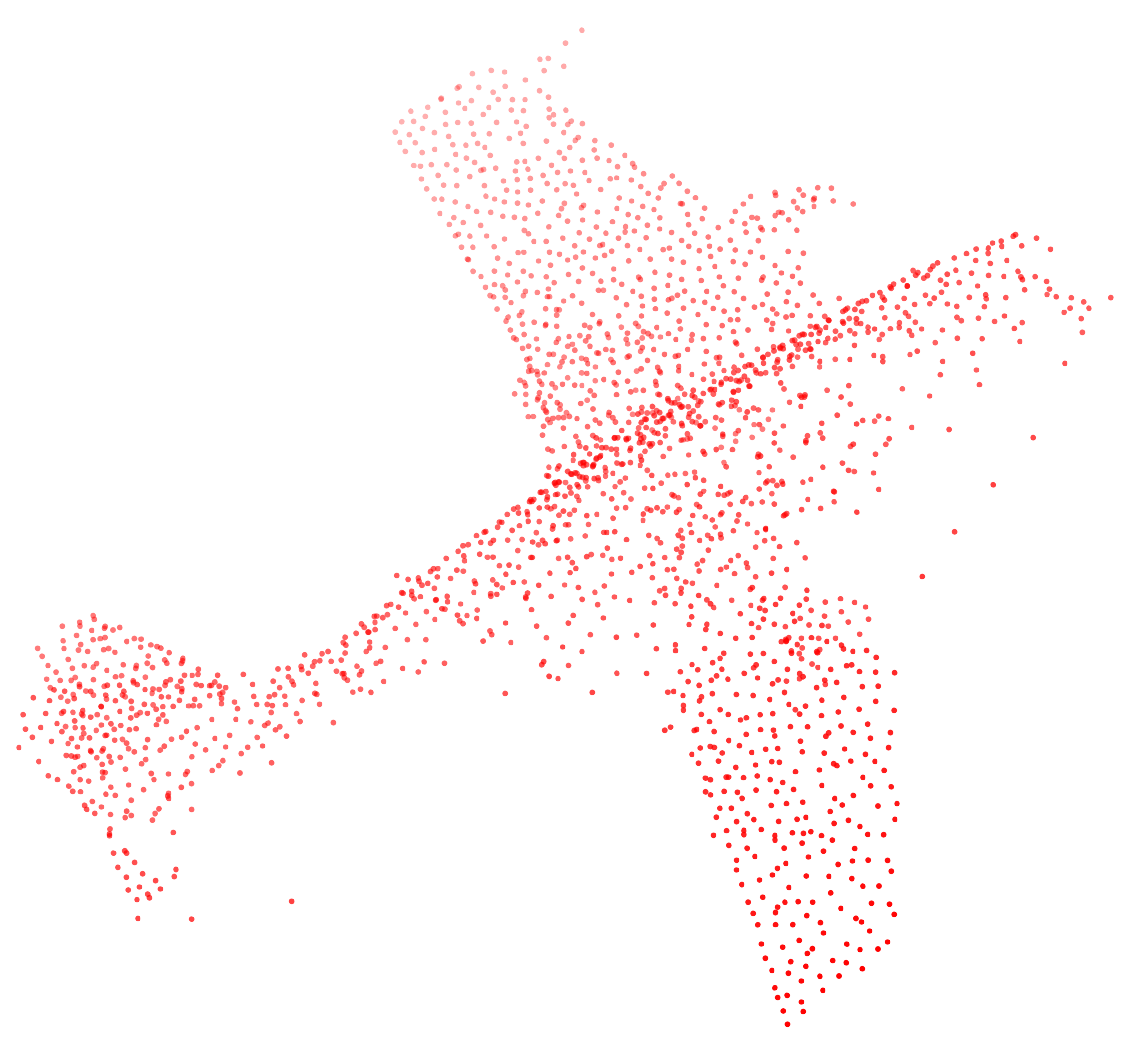}              
      & \includegraphics[width=0.1111\textwidth]{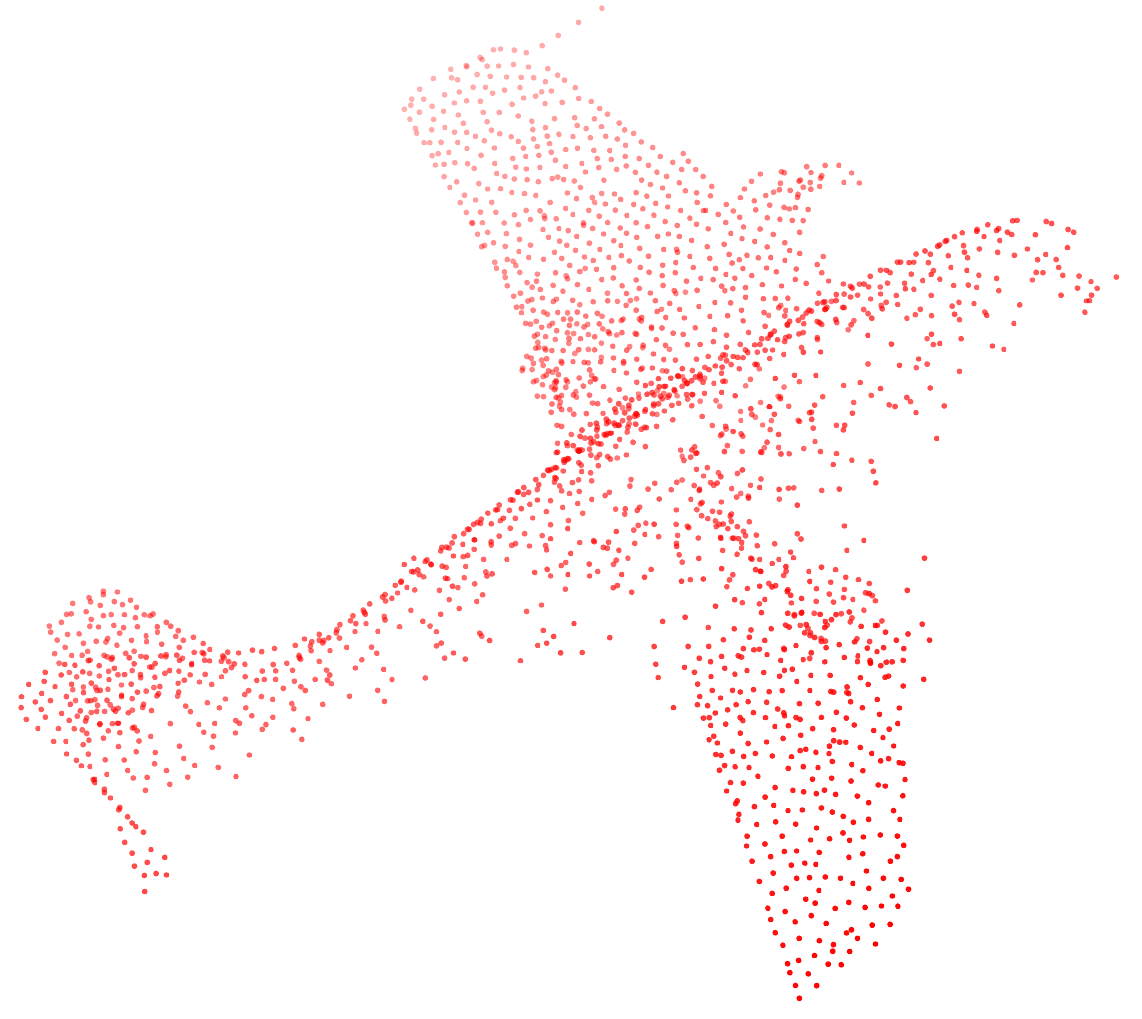}
      & \includegraphics[width=0.1111\textwidth]{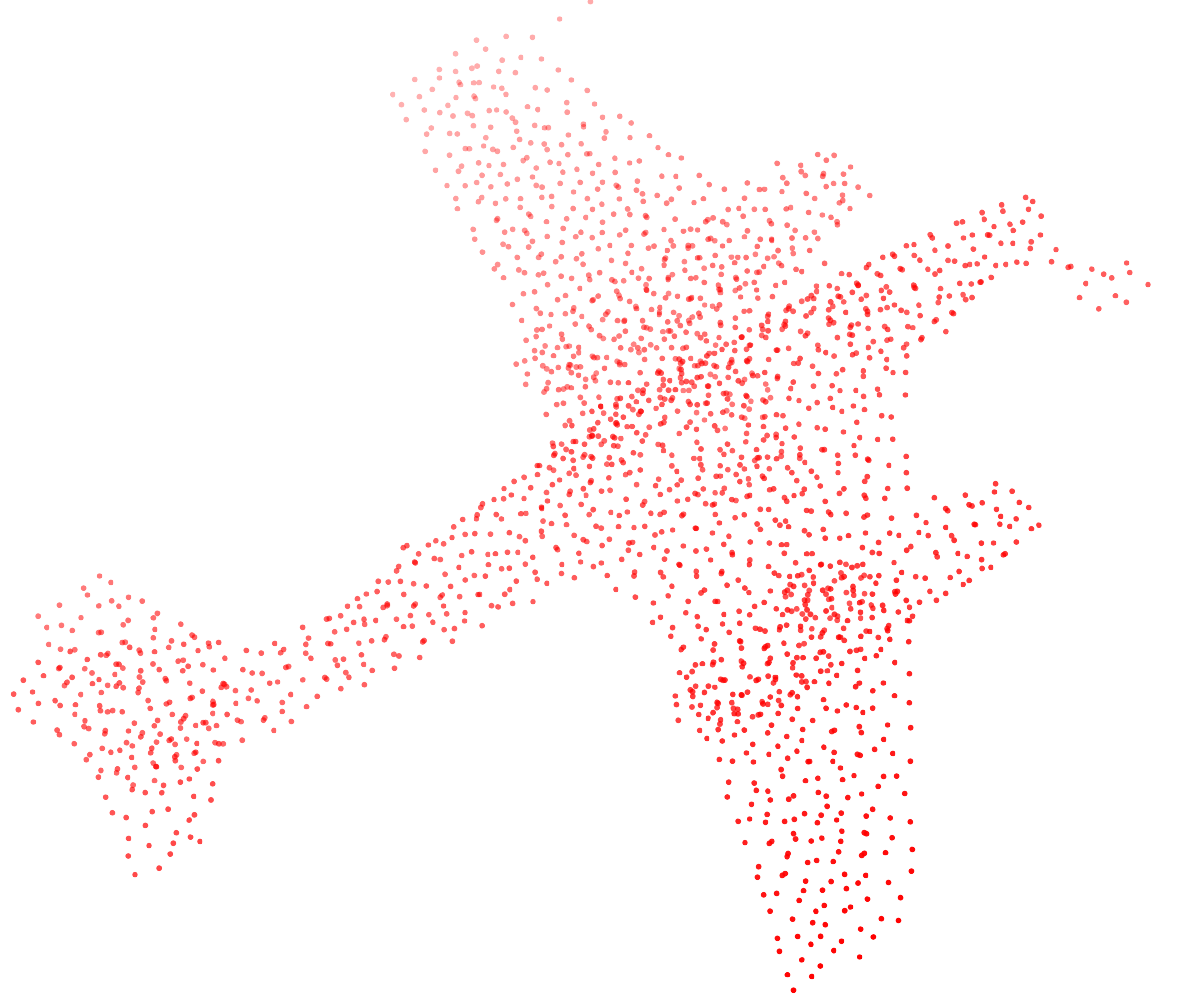} \\
      \specialrule{0em}{0pt}{5pt}
               \includegraphics[width=0.1111\textwidth]{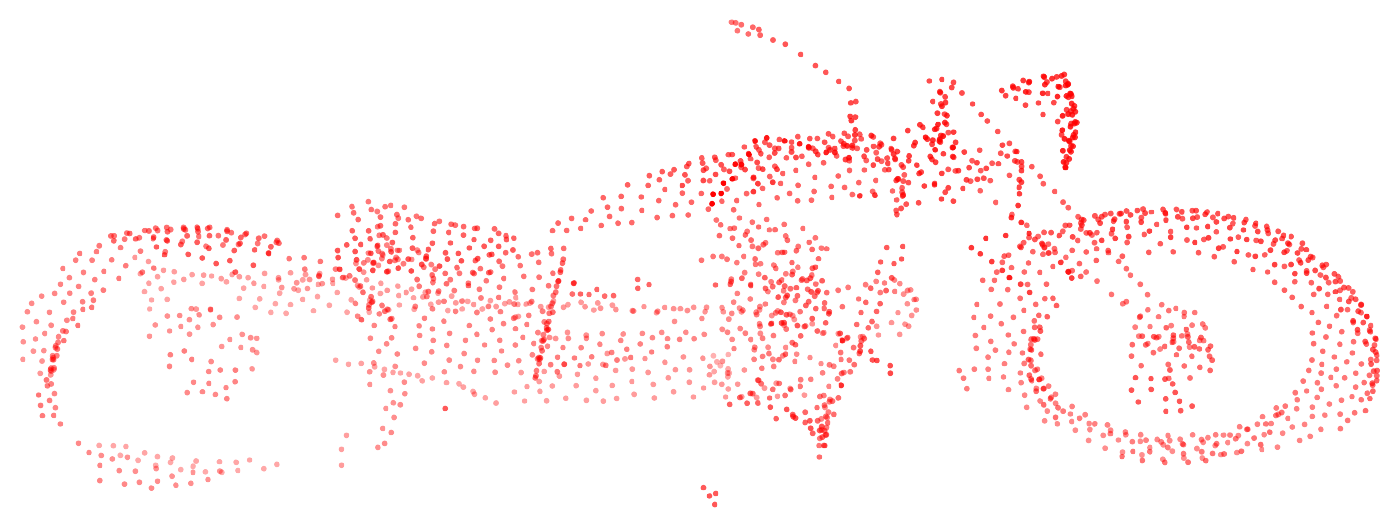}
      & \includegraphics[width=0.1111\textwidth]{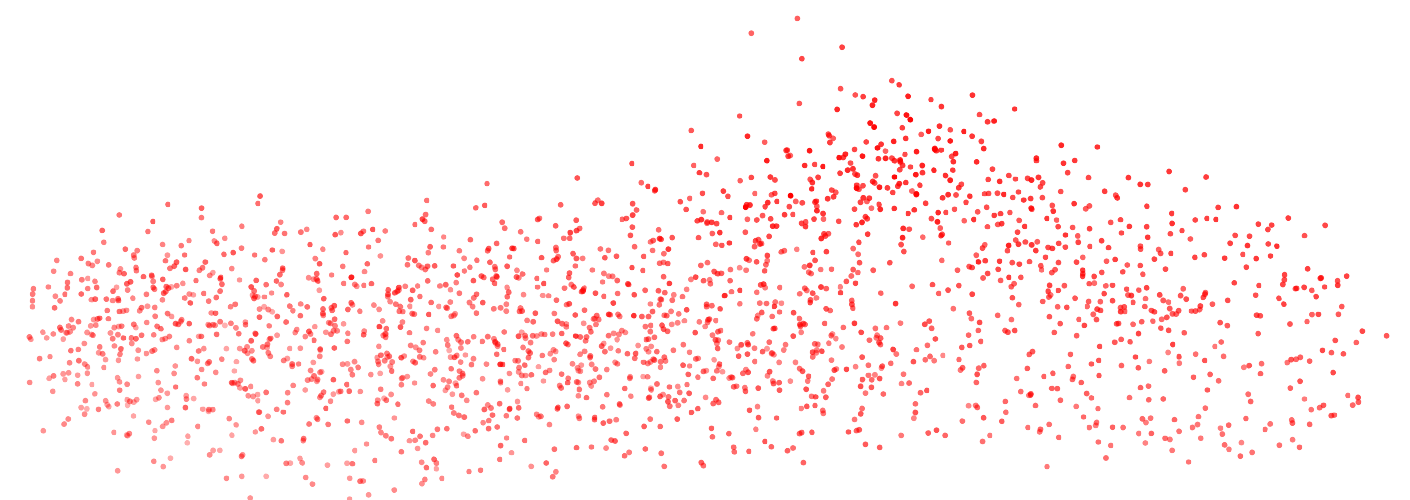}
      & \includegraphics[width=0.1111\textwidth]{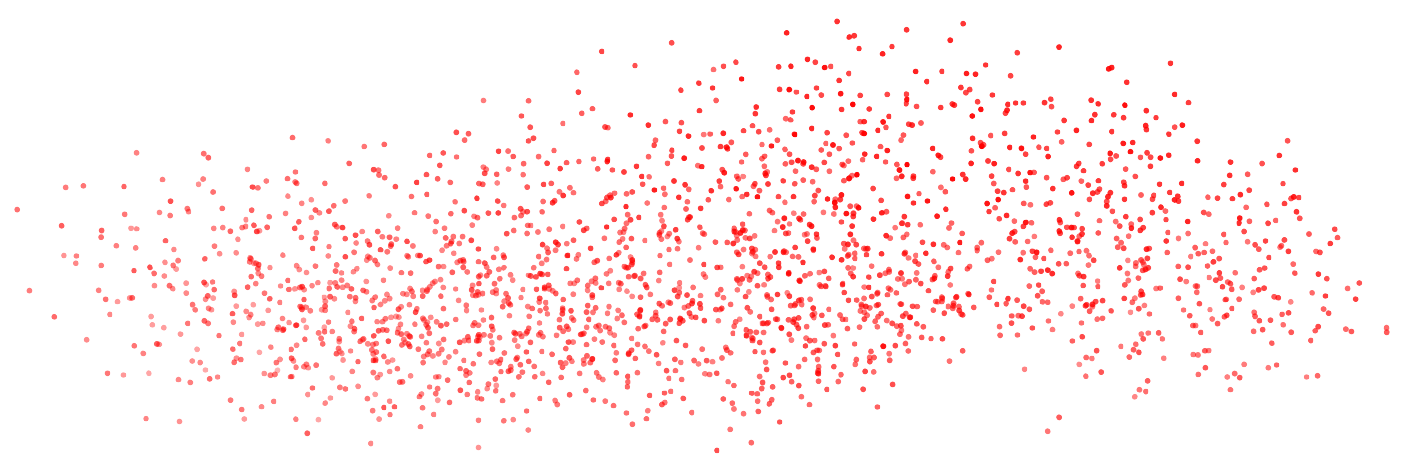}
      & \includegraphics[width=0.1111\textwidth]{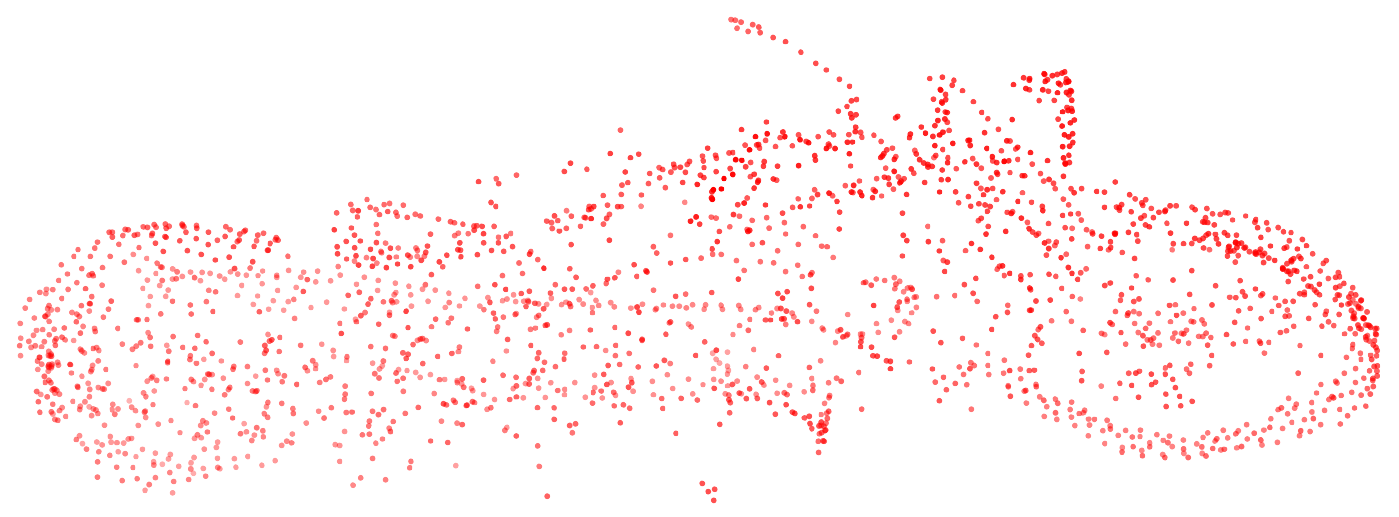}        
      & \includegraphics[width=0.1111\textwidth]{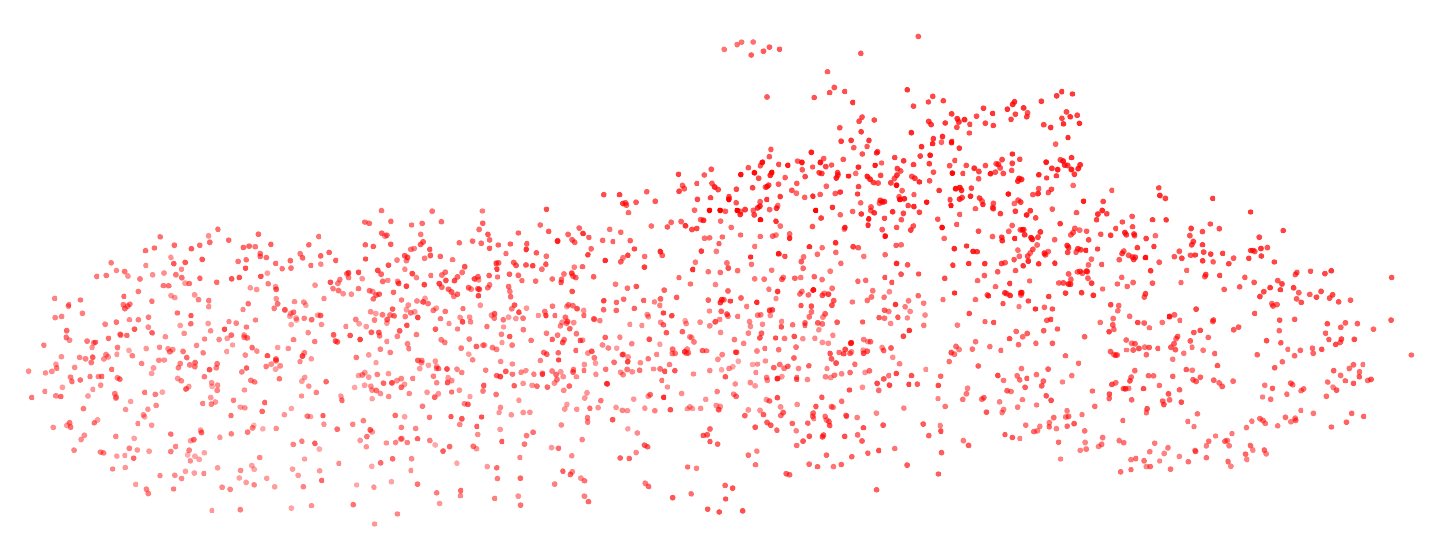}    
      & \includegraphics[width=0.1111\textwidth]{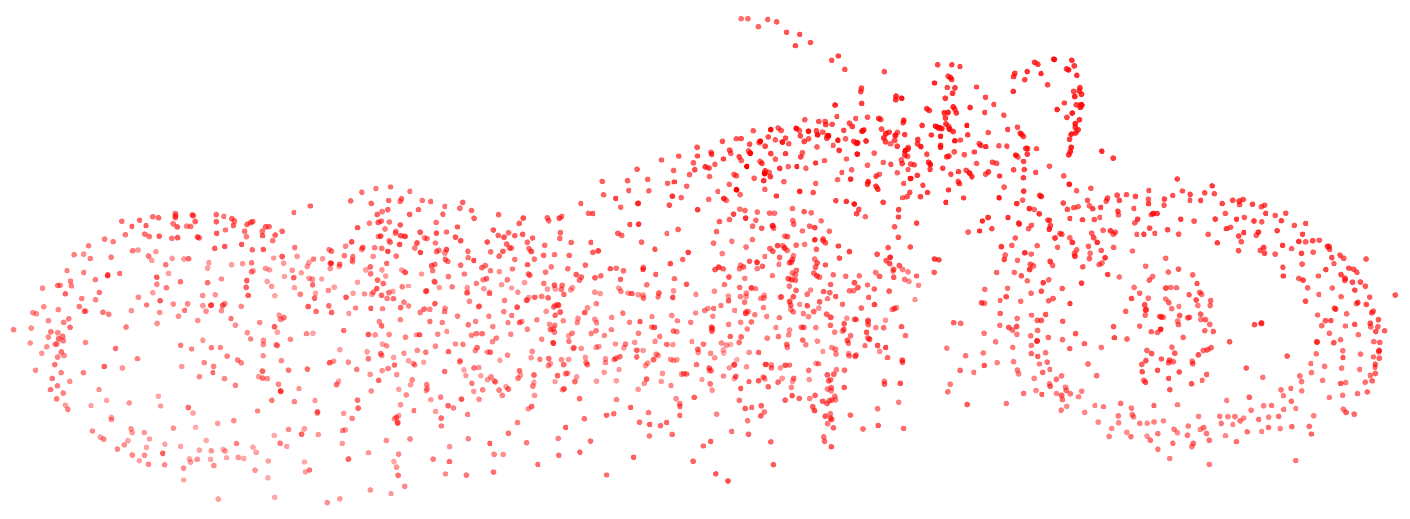}      
      & \includegraphics[width=0.1111\textwidth]{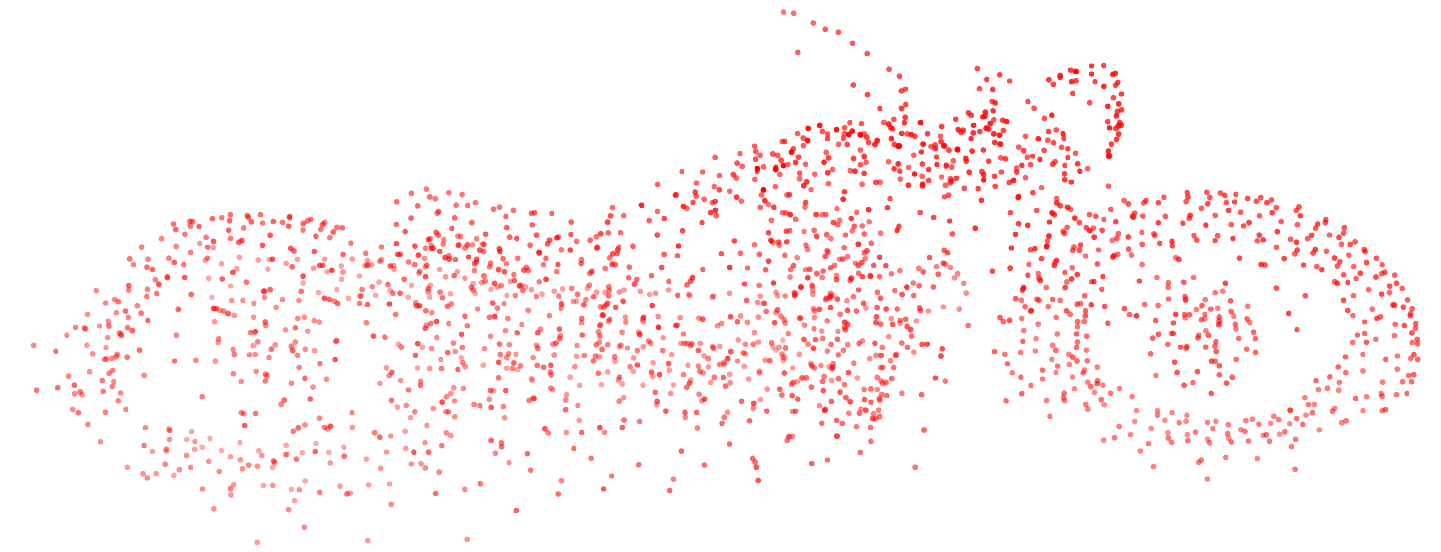}              
      & \includegraphics[width=0.1111\textwidth]{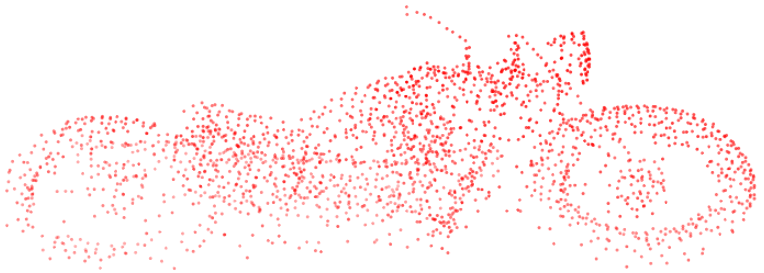}
      & \includegraphics[width=0.1111\textwidth]{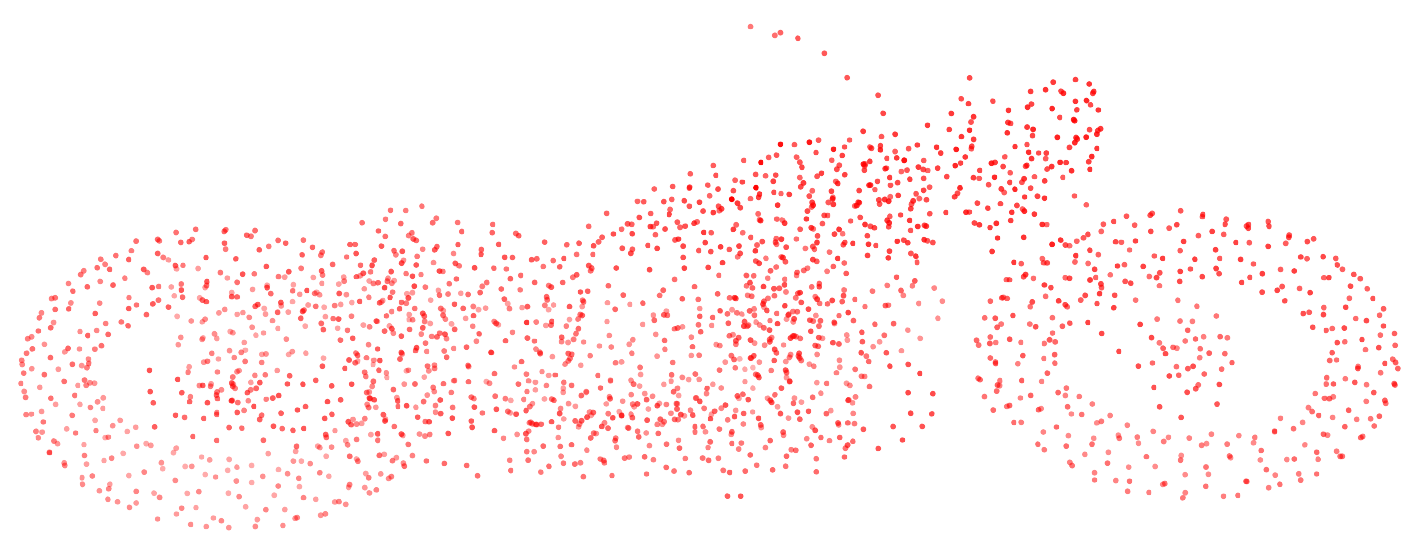} \\
      \specialrule{0em}{0pt}{5pt}
                    \includegraphics[width=0.1111\textwidth]{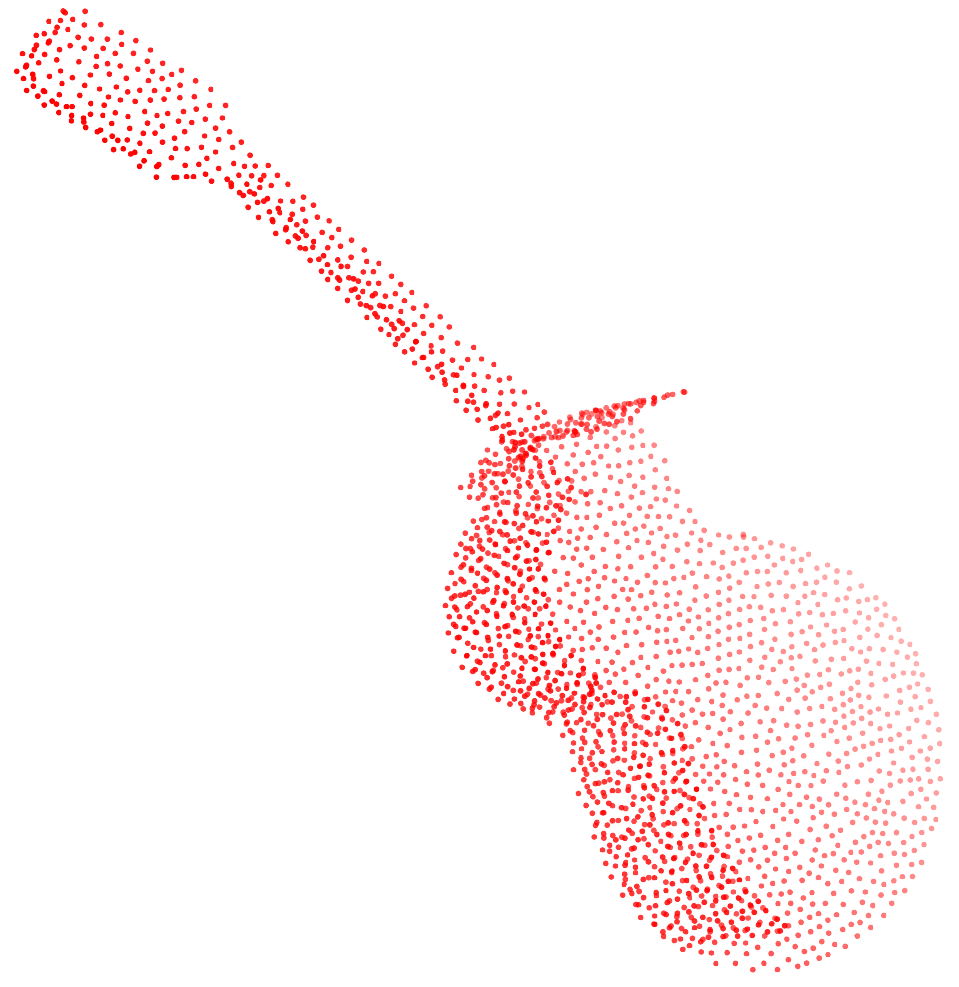}
      & \includegraphics[width=0.1111\textwidth]{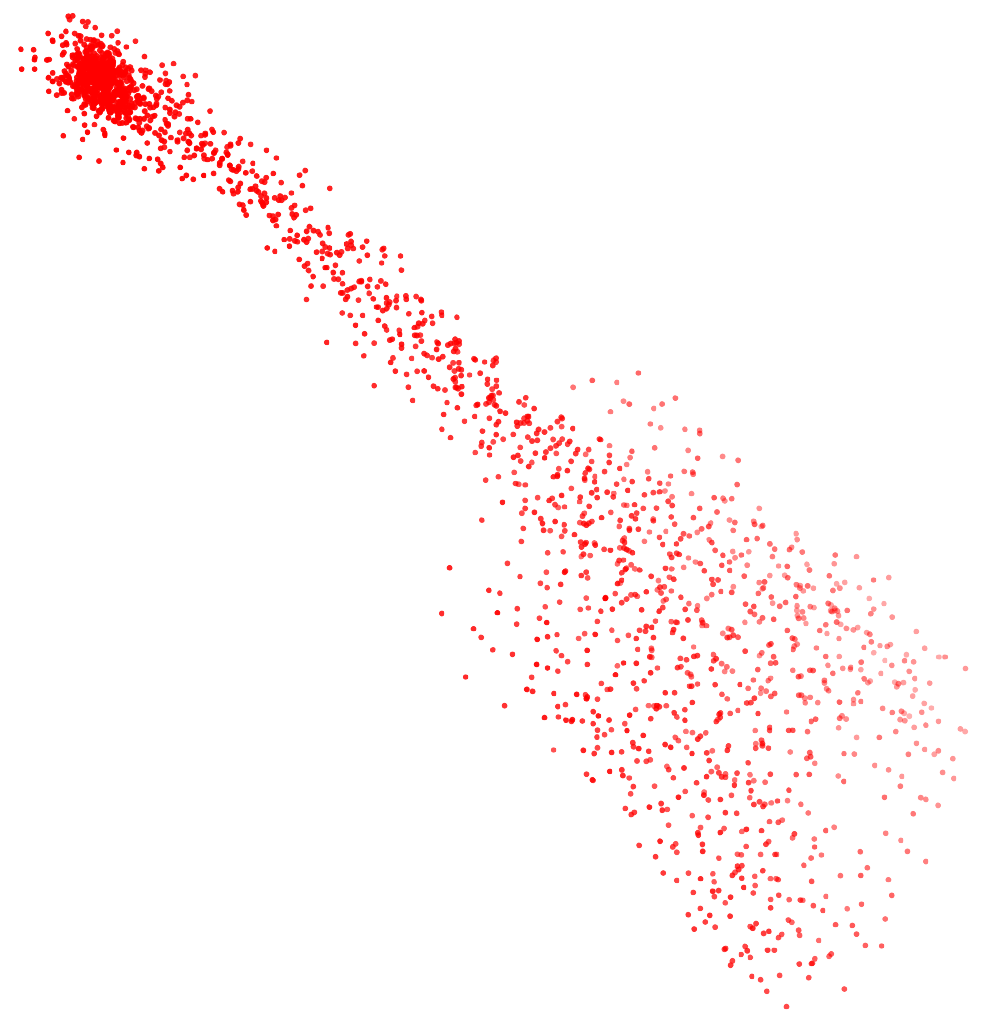}
      & \includegraphics[width=0.1111\textwidth]{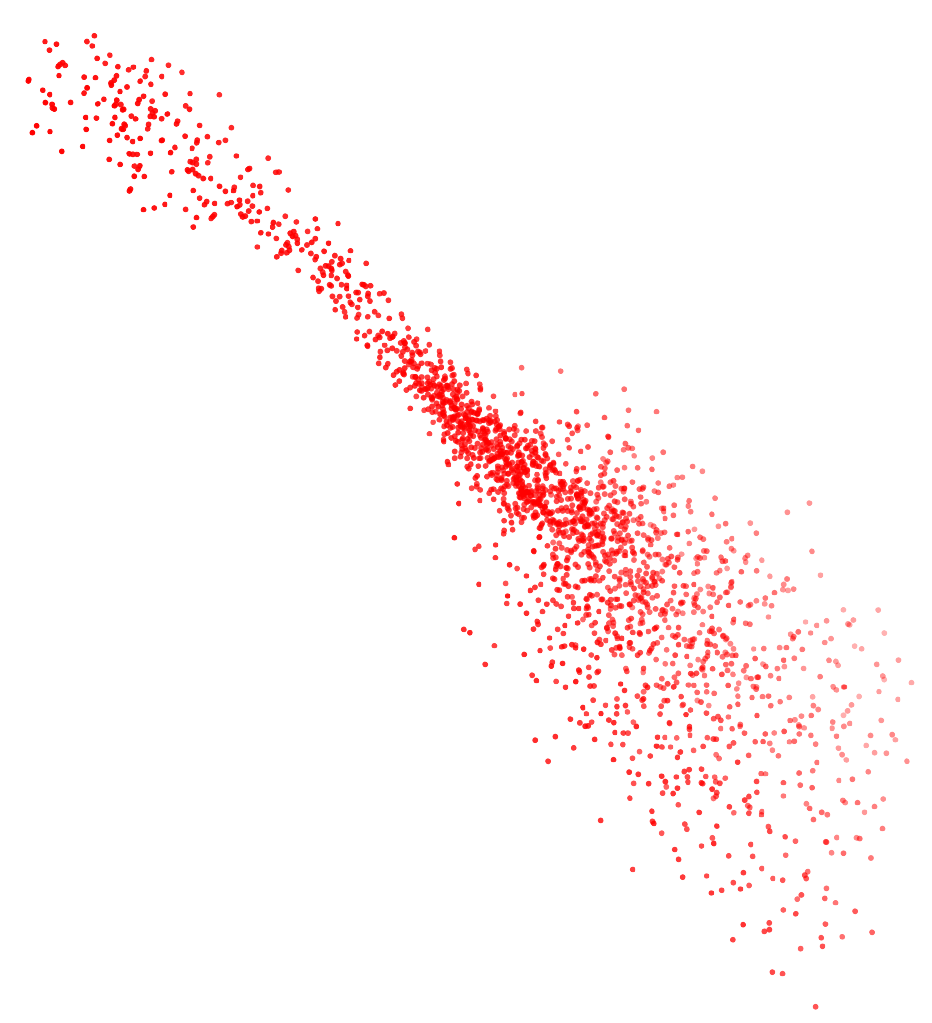}
      & \includegraphics[width=0.1111\textwidth]{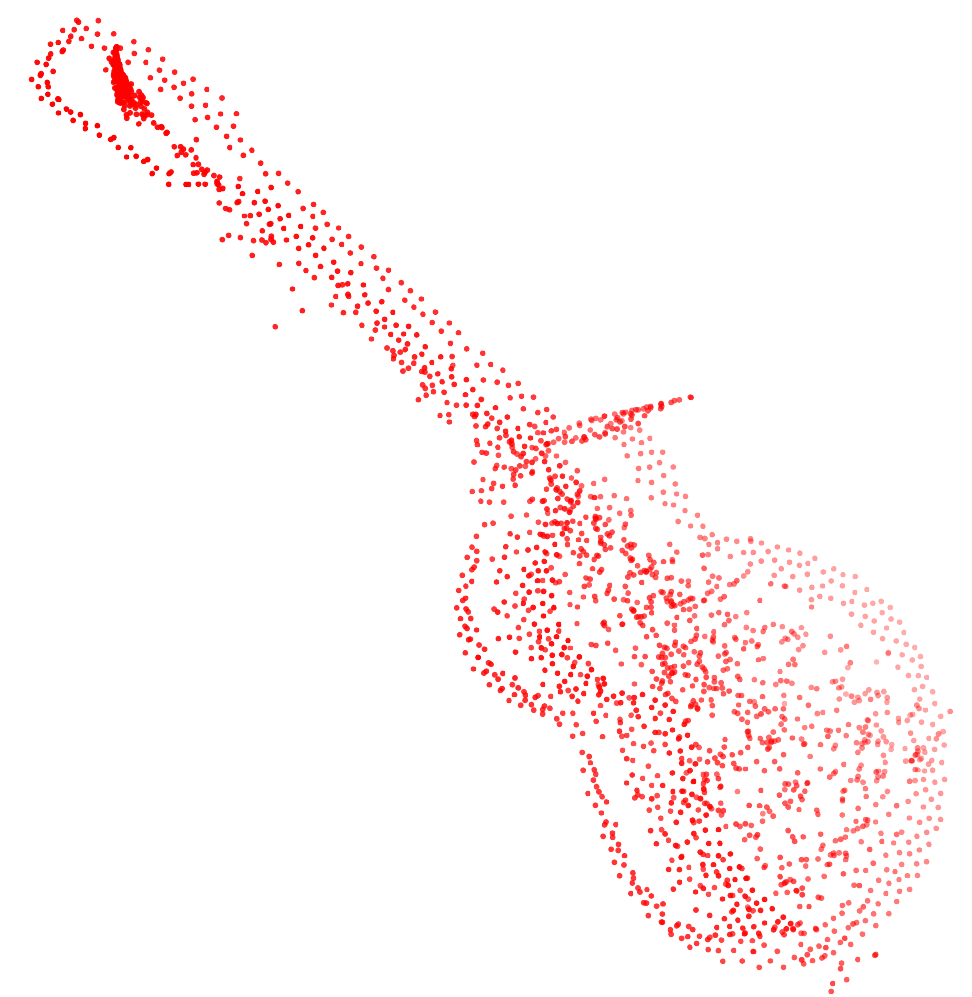}        
      & \includegraphics[width=0.1111\textwidth]{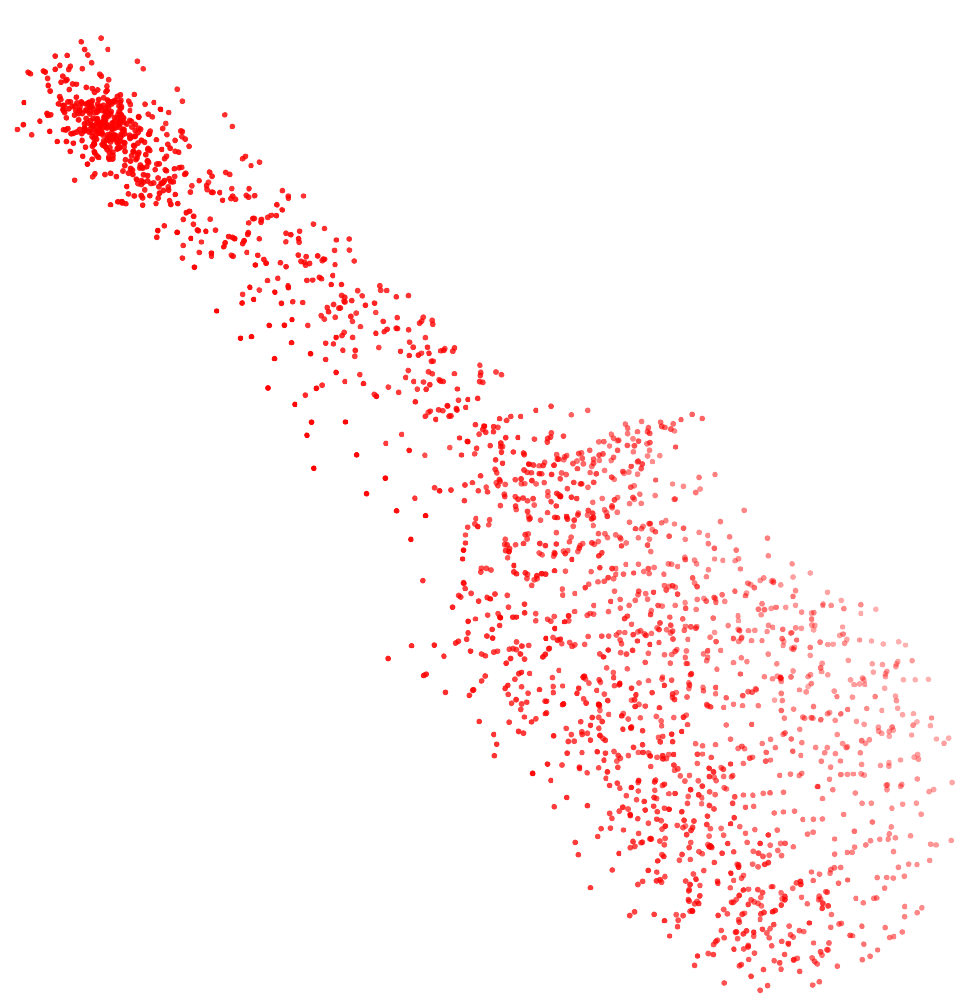}    
      & \includegraphics[width=0.1111\textwidth]{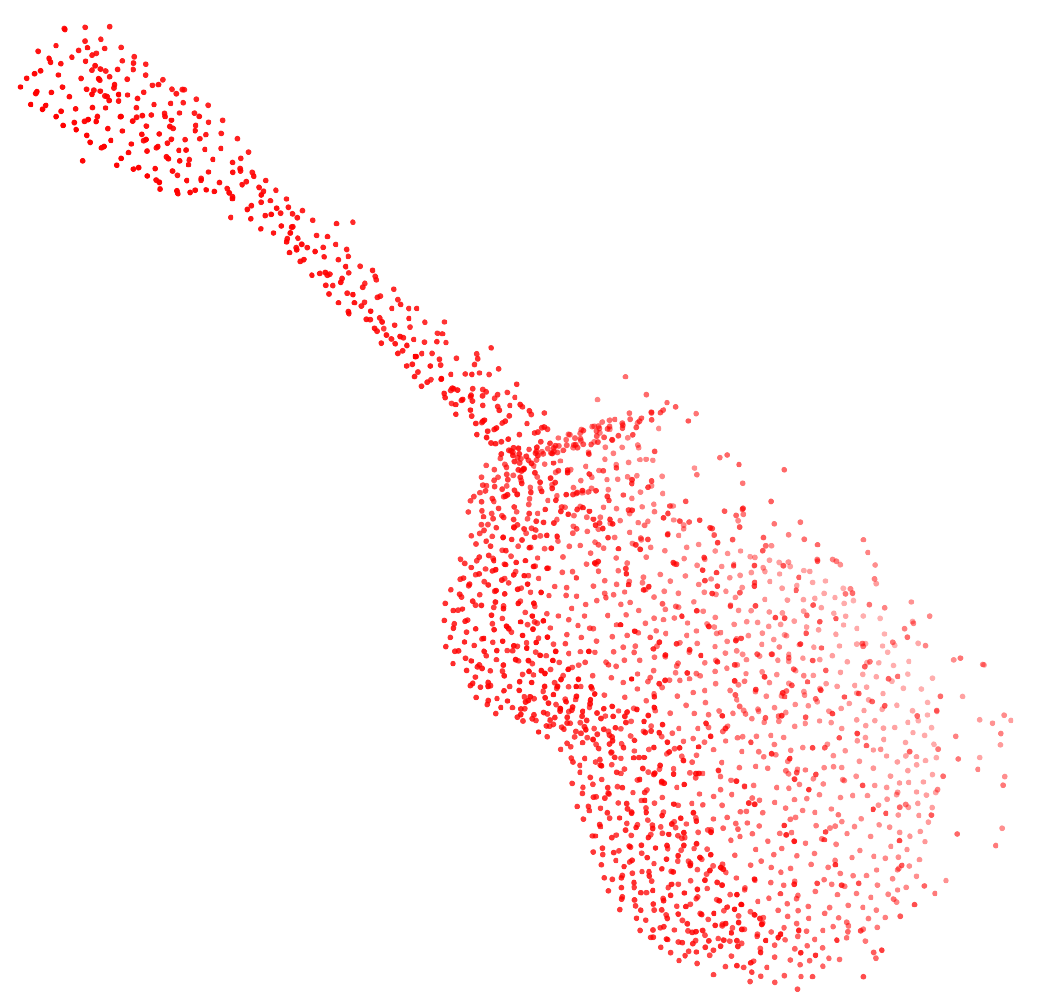}      
      & \includegraphics[width=0.1111\textwidth]{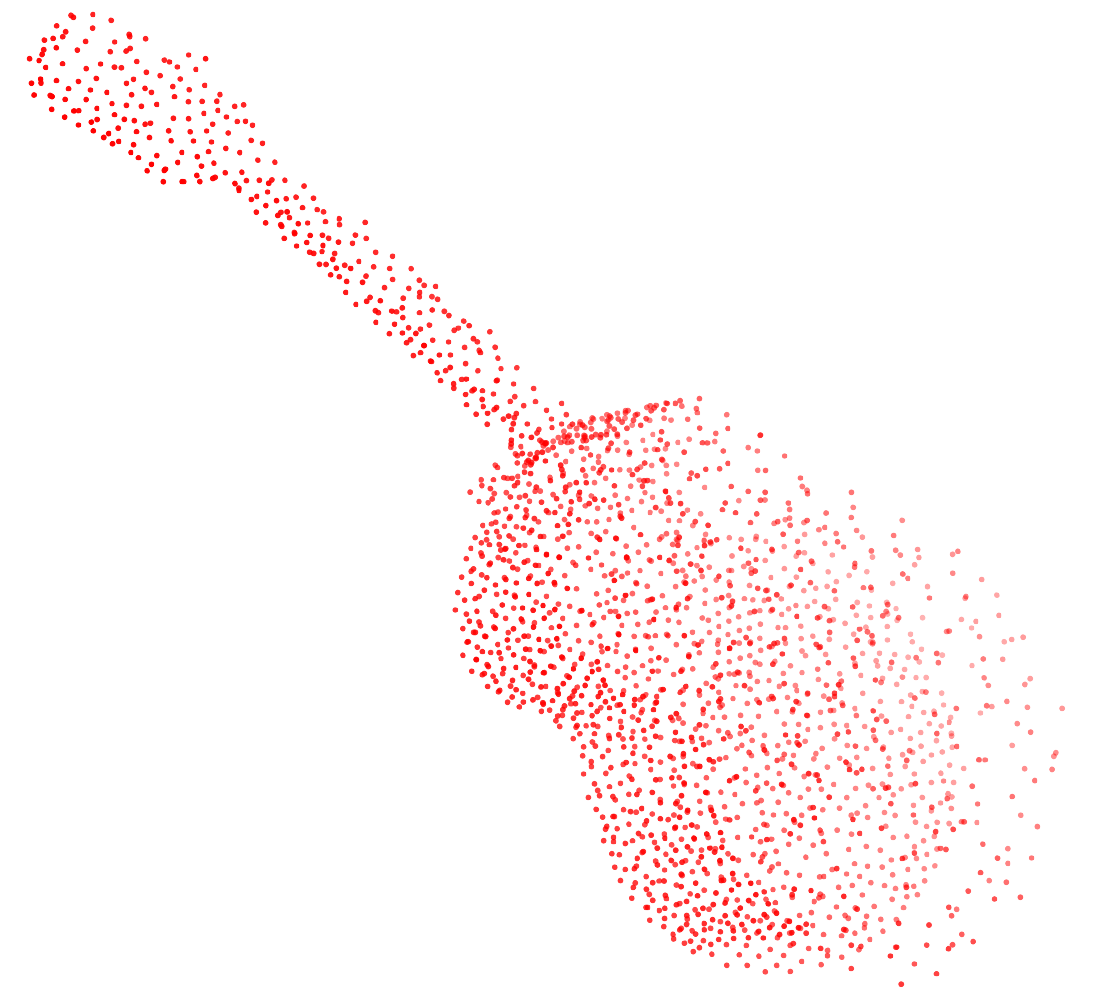}              
      & \includegraphics[width=0.1111\textwidth]{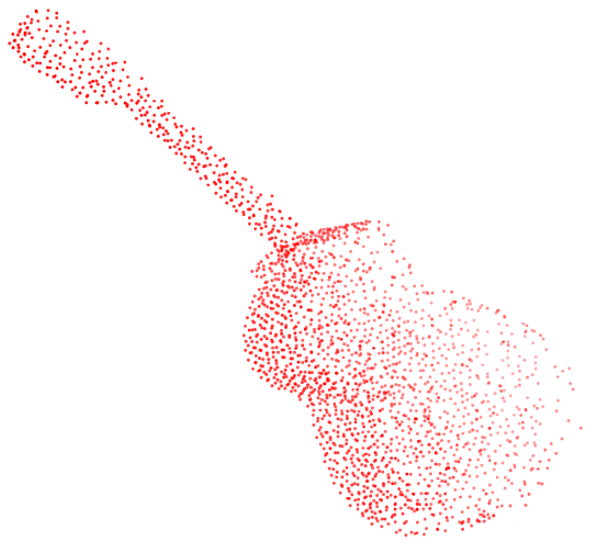}
      & \includegraphics[width=0.1111\textwidth]{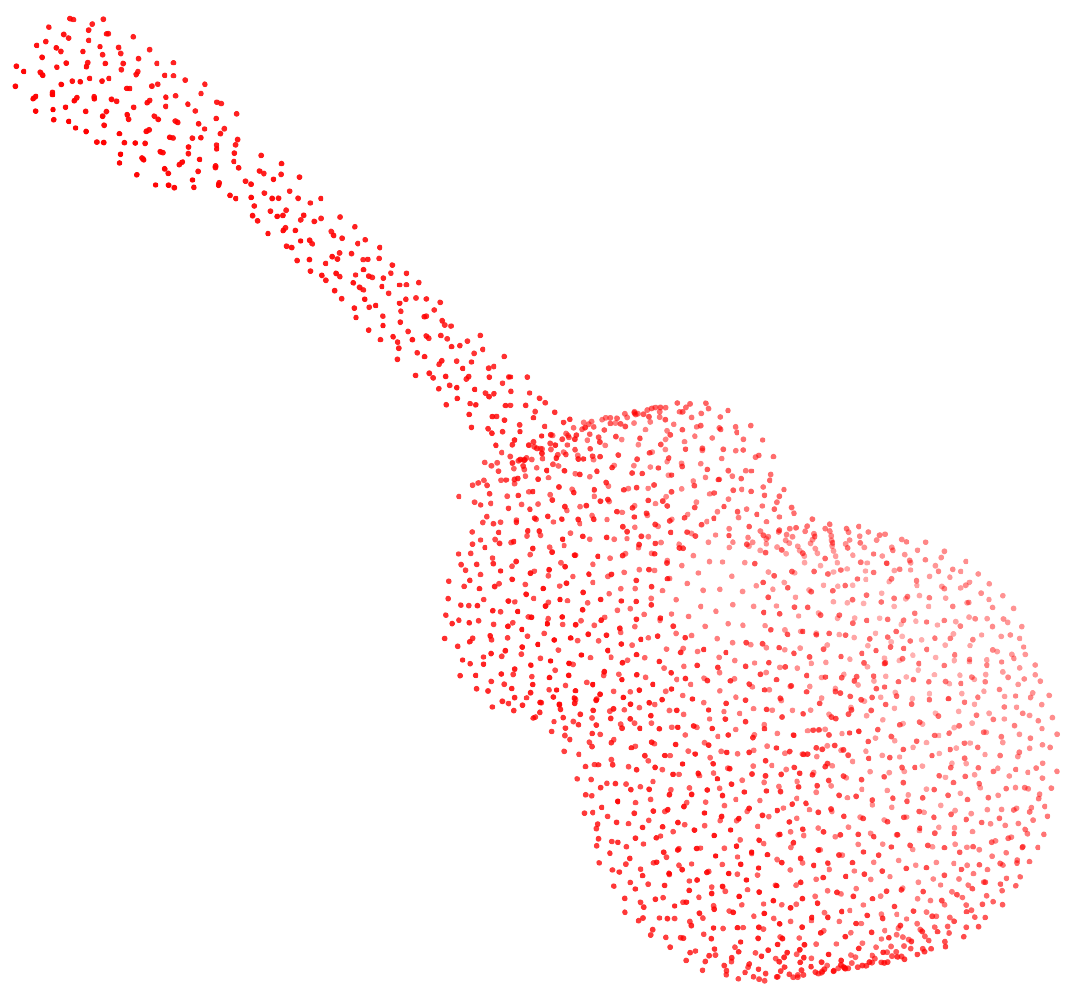} \\ 
\end{tabular}
}
\end{table}
\vspace{-0.5cm}
\vspace{-0.5cm}
\subsubsection{Qualitative Evaluation\label{Section 4.2}}
Table \ref{tab5} qualitatively shows the completion results on the transformed cloud by different methods. As we discussed in the previous section, \cite{2pcn,3,4ECG,5Cascade,6TopNet} neglect the fine-grained local details. Furthermore, unlike the completion results on fixed-pose point clouds in previous studies, we can clearly observe that existing methods exhibit a significant deterioration in completion performance on rotated point clouds. For example, TopNet \cite{6TopNet} almost  lacks the ability to complete local features on the rotated point cloud, and MSN \cite{3}  only roughly recovers the outline, leaving many intermediate points missing. RICNet generates complete shapes with finer details compared to other alternatives, with clearly observable relational structures. 

More specifically, compared with the results of VRCNet\cite{7}, RICNet achieves a more comprehensive completion of specific global characteristics. In the first row, RICNet produces chair completions with straighter legs. In the second row, VRCNet shows a higher occurrence of outliers in the completion of airplanes, while RICNet demonstrates a cleaner and more precise representation of the surface. In the third row,  VRCNet neglects the completion of the missing motorcycle headlight, whereas RICNet exhibits remarkable proficiency in recovering the overall point cloud structure of motorcycles, highlighting its enhanced completion capabilities for local features. Lastly, in the fourth row, RICNet accurately recovers the smooth and rounded edge contours of guitars, preserving intricate details and ensuring faithful reconstruction. These achievements are attributed to our robust rotation-invariant point cloud completion network, which effectively handles the completion of randomly rotated point clouds.
\vspace{-0.5cm}

\subsubsection{Ablation Study\label{Section 4.3}}
This section presents the results of the ablation study conducted on RICNet to evaluate the effectiveness of each component. The study focuses on three key components of our model: the rotation-invariant encoder, the dual-path architecture, and the enhancing module. We use PCN\cite{2pcn} to denote the model that does not incorporate any of the three aforementioned modules. Table \ref{tab6} illustrates the results of the ablation study conducted on our proposed modules. The completion results are evaluated with 2048 points. The results of the ablation study convincingly highlight the crucial significance of the proposed module within our network.
\vspace{-0.5cm}
\begin{table}[t]
\renewcommand\arraystretch{1.25}
\centering
\caption{Ablation experiments are conducted on the proposed network modules. These experiments examine the effectiveness of rotation-invariant encoder, dual-path architecture, and the enhancing module.} \label{tab6}
\scalebox{0.8}{
\begin{tabular}{ccc|cc}
\hline
\begin{tabular}[c]{@{}c@{}}Enhancing\\ Module\end{tabular} & \begin{tabular}[c]{@{}c@{}}Dual-Path\\ Architecture\end{tabular} & \begin{tabular}[c]{@{}c@{}}Rotation-Invariant\\ Encoder\end{tabular} & CD$\downarrow$ & F1$\uparrow$\\ \hline
          &                    &                      &15.68                  &0.236    \\
\(\surd\)          &                    &                      &8.51                  &0.468    \\
\(\surd\)          &\(\surd\)                    &                      &7.75                  &0.469    \\
 \(\surd\)         &\(\surd\)                    &\(\surd\)                       &$\textbf{7.57}$                 &$\textbf{0.477}$    \\ \hline
\end{tabular}
}
\end{table}
\section{Conclusion}
In this paper, we propose RICNet, a rotation-invariant point cloud completion network. Our feature extraction module exhibits robustness to rigid transformations such as rotation and translation of the input point clouds. This module can be transferred to other point cloud tasks to enhance the network's rotation invariance. We evaluate the completion performance of existing methods and our network on rotated point clouds through comprehensive experiments. RICNet exhibits excellent generalization performance on incomplete point clouds with different poses, making it suitable for practical applications like robotics, autonomous driving, and 3D reconstruction. RICNet benefits downstream tasks like point cloud segmentation, classification, and registration. Our network modules are also available for future research on incomplete point clouds.

%
%
%

\begin{thebibliography}{8}
\bibitem{1pointnet}
C. R. Qi, H. Su, K. Mo, et al.: PointNet: Deep learning on point sets for 3D classification and segmentation. In: Proceedings of the IEEE Conference on Computer Vision and Pattern Recognition. pp. 652–660 (2017)
\bibitem{2pcn}
Wentao Yuan, Tejas Khot, David Held, et al.: Pcn: Point completion network. In: International Conference on 3D Vision (3DV). pp. 728–737 (2018)
\bibitem{3}
Minghua Liu, Lu Sheng, Sheng Yang, et al.: Morphing and sampling network for dense point cloud completion. In: Proceedings of the AAAI Conference on Artificial Intelligence. pp. 11596–11603 (2020)
\bibitem{4ECG}
Liang Pan.: Ecg: Edge-aware point cloud completion with graph convolution. In: IEEE Robotics and Automation Letters (2020)
\bibitem{5Cascade}
Xiaogang Wang, Marcelo H Ang Jr, and Gim Hee Lee.: Cascaded refinement network for point cloud completion. In: Proceedings of the IEEE/CVF Conference on Computer Vision and Pattern Recognition. pp. 790–799 (2020)
\bibitem{6TopNet}
Lyne P Tchapmi, Vineet Kosaraju, Hamid Rezatofighi, et al.: Topnet: Structural point cloud decoder. In: Proceedings of the IEEE Conference on Computer Vision and Pattern Recognition. pp. 383–392 (2019)
\bibitem{7}
L. Pan, X. Chen, Z. Cai, et al.: Variational relational point completion network. In: Proceedings of the IEEE/CVF Conference on Computer Vision and Pattern Recognition. pp. 8524-8533 (2022)
\bibitem{8}
Chuanxia Zheng, Tat-Jen Cham, and Jianfei Cai.: Pluralistic image completion. In: Proceedings of the IEEE Conference on Computer Vision and Pattern Recognition. pp. 1438–1447 (2019)
\bibitem{9}
Qi, C.R., Yi, L., Su, H., et al.: Pointnet++: Deep hierarchical feature learning on point sets in a metric space. In: Advances in Neural Information Processing Systems. pp. 5105–5114 (2017)
\bibitem{10}
Tombari, F., Salti, et al.: Unique signatures of histograms for local surface description. In: Proceedings of the European Conference on Computer Vision. pp. 356–369 (2010)
\bibitem{11}
Zhang Z, Hua B S, Rosen D W, et al.: Rotation invariant convolutions for 3d point clouds deep learning. In: International Conference on 3D Vision. pp. 204–213 (2019)
\bibitem{12}
Zhang Z, Hua B S, Chen W, et al.: Global context aware convolutions for 3d point cloud understanding. In: International Conference on 3D Vision. pp. 210-219 (2020)
\bibitem{13}
Kim, S., Park, J., et al.: Rotation-invariant local-to-global representation learning for 3d point cloud. In: Advances in Neural Information Processing Systems. pp. 8174–8185 (2020)
\bibitem{14}
Thomas, H.: Rotation-invariant point convolution with multiple equivariant alignments. In: 2020 International Conference on 3D Vision (3DV). pp. 504-513 (2020)
\bibitem{15}
Li, X., Li, et al.: A rotation-invariant framework for deep point cloud analysis. IEEE Transactions on Visualization and Computer Graphics, 4503-4514 (2021)
\bibitem{16}
Zhang, Z., Hua, S.K.: RIConv++: Effective Rotation Invariant Convolutions for 3D Point Clouds Deep Learning. International Journal of Computer Vision, 1228-1243 (2022)
\bibitem{17}
Wang, Y., Sun, et al.: Dynamic graph cnn for learning on point clouds. ACM Transactions on Graphics, 1-12 (2019)
\bibitem{18}
Arno Knapitsch, Jaesik Park, Qian-Yi Zhou, et al.: Tanks and temples: Benchmarking large-scale scene reconstruction. ACM Transactions on Graphics (ToG), 1–13 (2017)
\bibitem{19}
Tatarchenko M, Richter S R, Ranftl R, et al.: What do single-view 3d reconstruction networks learn? In: Proceedings of the IEEE/CVF conference on computer vision and pattern recognition. pp. 3405-3414 (2019)
\bibitem{20}
Chen B, Fan J, Zhao P, et al.: Slice Sequential Network: A Lightweight Unsupervised Point Cloud Completion Network. In: Pattern Recognition and Computer Vision: 4th Chinese Conference, PRCV 2021. pp. 103-114 (2021)
\bibitem{21}
Tao M, Zhao C, Wang J, et al.: Global Patch Cross-Attention for Point Cloud Analysis. In: Pattern Recognition and Computer Vision: 5th Chinese Conference, PRCV 2022. pp. 96-111 (2022)
\bibitem{22}
Pan L, Chen X, Cai Z, et al.: Variational Relational Point Completion Network for Robust 3D Classification. IEEE Transactions on Pattern Analysis and Machine Intelligence (2023)
\end{thebibliography}
%

\end{document}